%% file: main.tex
\documentclass[11pt,a4paper]{article}
\usepackage{times,latexsym}
\usepackage{url}
\usepackage[T1]{fontenc}

\usepackage[acceptedWithA]{tacl2021v1}

\usepackage{xspace,mfirstuc,tabulary}
\usepackage[caption=false]{subfig}

\newif\iftaclinstructions
\taclinstructionsfalse %
\iftaclinstructions
\renewcommand{\confidential}{}
\renewcommand{\anonsubtext}{(No author info supplied here, for consistency with
TACL-submission anonymization requirements)}
\newcommand{\instr}
\fi

\iftaclpubformat %

\else

\fi

\input{preamble}

\title{\dataset: Evaluating Question Answering on Debatable Knowledge}

\author{Rongwu Xu$^{1*}$, Xuan Qi$^{1*}$, \\ \bf Zehan Qi$^{1}$, Wei Xu$^{1}$, Zhijiang Guo$^{2}$ \\
$^{1}$ Tsinghua Universty, $^{2}$ University of Cambridge\\
\texttt{Emails: \{xrw22, qi-x22@\}@mails.tsinghua.edu.cn}\\
$^{*}$ Equal contribution
}

\date{}

\begin{document}
\maketitle
\begin{abstract}
The rise of large language models (LLMs) has enabled us to seek answers to inherently debatable questions on LLM chatbots, necessitating a reliable way to evaluate their ability. However, traditional QA benchmarks assume fixed answers are inadequate for this purpose. To address this, we introduce \dataset{}, a dataset of 2,941 debatable questions, each accompanied by multiple human-annotated partial answers that capture a variety of perspectives.
We develop two metrics: Perspective Diversity, which evaluates the comprehensiveness of perspectives, and Dispute Awareness, which assesses if the LLM acknowledges the question's debatable nature. Experiments demonstrate that both metrics are aligned with human preferences and stable across different underlying models.
Using \dataset{} with two metrics, we assess 12 popular LLMs and retrieval-augmented generation methods. 
Our findings reveal that while LLMs generally excel at recognizing debatable issues, their ability to provide comprehensive answers encompassing diverse perspectives varies considerably.
\end{abstract}

\vspace{-2em}
\noindent
\begin{wrapfigure}{l}{0.05\textwidth}
    \centering
    \hypertarget{github-link}{}
    \href{https://github.com/pillowsofwind/DebateQA}{%
    \includegraphics[width=0.05\textwidth]{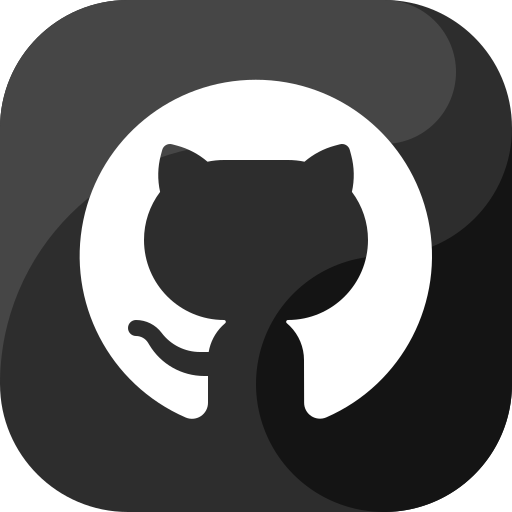}
    }
\vspace{-2em}
\end{wrapfigure}

\noindent
{\fontsize{10}{10}\selectfont\url{https://github.com/pillowsofwind/DebateQA}}

\input{Secs/intro}
\input{Tabs/dataset_sample}
\input{Secs/related_work}
\input{Secs/dataset}
\input{Secs/criteria}

\input{Secs/metric_PD}
\input{Secs/metric_DA}

\input{Secs/experiments}
\input{Secs/conclusion}

\clearpage
\bibliography{custom}
\bibliographystyle{acl_natbib}

\clearpage

\appendix

\input{Appens/dataset}

\input{Appens/metric_PD}

\input{Appens/metric_DA}

\input{Appens/experiments}

\input{Appens/case_study}

\clearpage
\onecolumn
\input{Appens/interpretation}

\end{document}

%% file: preamble.tex
\usepackage{amsmath}
\usepackage{amssymb}
\usepackage{mathtools}

\usepackage{tikz}
\usepackage{soul}
\usepackage{xcolor}

\usepackage{booktabs}
\usepackage{threeparttable}
\usepackage{tabularx}
\usepackage{multirow}
\usepackage{multicol}

\usepackage{inconsolata}

\usepackage{algorithmic}
\usepackage[linesnumbered,ruled,vlined]{algorithm2e}
\SetKwInput{KwData}{Input}
\SetKwInput{KwResult}{Output}
\SetKwComment{tcc}{$\#$}{}

\SetCommentSty{mycommfont}

\usepackage{wrapfig}

\newcommand{\ctext}[2]{%
    \tikz[baseline=(X.base)] \node[fill=#1,rounded corners=1.8pt,inner sep=1pt] (X) {#2};%
}

\newcommand{\ie}{{\em i.e.}}
\newcommand{\eg}{{\em e.g.}}

\newcommand{\wrt}{{\em w.r.t.}}

\definecolor{mygreen}{RGB}{11,141,10}
\definecolor{myred}{RGB}{223,68,52}
\definecolor{myblue}{RGB}{70,130,180}
\definecolor{mydeepblue}{RGB}{65,105,225}
\definecolor{myviolet}{RGB}{97,0,138}
\definecolor{myburgundy}{RGB}{110,10,30}
\definecolor{myblue2}{RGB}{0,105,148}
\definecolor{iceblue}{RGB}{173, 216, 230}
\definecolor{puregreen}{RGB}{0, 70, 0}

\definecolor{sourcequestions}{RGB}{194, 114, 68}
\definecolor{collectingpartialanswer}{RGB}{193, 155, 68}
\definecolor{humanannotation}{RGB}{111, 141, 88}

\newenvironment{packeditemize}{
\begin{list}{$\bullet$}{
\setlength{\labelwidth}{8pt}
\setlength{\itemsep}{0pt}
\setlength{\leftmargin}{\labelwidth}
\addtolength{\leftmargin}{\labelsep}
\setlength{\parindent}{0pt}
\setlength{\listparindent}{\parindent}
\setlength{\parsep}{3pt}
\setlength{\topsep}{3pt}}}{\end{list}}

\usepackage{pifont}
\newcommand{\crossmark}{\textcolor{myred}{\text{\ding{55}}}}
\renewcommand{\checkmark}{\textcolor{mygreen}{\text{\ding{51}}}}

\newcommand{\dataset}{\textsc{DebateQA}}

\newcommand{\pov}{POV}
\newcommand{\explan}{Explan}

\newcommand{\placeholder}[1]{\textcolor{blue!80}{\bf\{}{#1}\textcolor{blue!80}{\bf\}}}

\newcommand{\criteriaone}{Perspective Diversity (P.D.)}
\newcommand{\criteriatwo}{Dispute Awareness (D.A.)}

\newcommand{\llmeval}{$\mathcal{M}_\text{eval}$}

%% file: Secs/intro.tex
\section{Introduction}
\label{sec: introduction}

How often do you query a chatbot about a debatable issue? Questions like ``Does Donald Trump have a terrible character?'' or ``How do crop circles form?'' frequently arise in everyday life, reflecting human beings' natural curiosity about topics that inherently lack fixed answers~\cite{lowry1981effects, brady2009curiosity}. With the advent of large language models (LLMs;~\citealt{achiam2023gpt, chowdhery2023palm, touvron2023llama}), we now turn to these models to seek ``\emph{proper}'' answers to such questions. Evaluating the performance of language models on debatable queries is crucial for understanding and enhancing their capabilities. 

Inherent difficulties hinder our way of assessing these models. Traditional question-answering (QA) benchmarks are typically designed to provide fixed answers to questions, as in datasets like SQuAD~\cite{rajpurkar2016squad} and Natural Questions~\cite{kwiatkowski2019natural}. Even in scenarios where multiple legitimate answers are possible, such as TruthfulQA~\cite{lin2022truthfulqa}, the answers are often presented as multiple-choice questions,
thus limiting the space of responses. Recent years have seen the emergence of long-form QA evaluations, such as ELI5~\cite{fan2019eli5} and ASQA~\cite{stelmakh2022asqa}, which allow for more elaborate answers. However, these works have not focused on inherently debatable questions.
The most related work to us is DELPHI~\cite{DBLP:conf/emnlp/SunAKKXW23}, which curates a dataset with controversial questions, however, their emphasis is on identifying controversy rather than delving deeper into evaluating models' responses.

\emph{What constitutes a proper answer when asked about debatable knowledge?} Drawing inspiration from interdisciplinary literature, we propose two desirable properties for such answers. Firstly, the answer should inform the receiver that the issue at hand is debatable~\cite{misco2011teaching}. Secondly, a proper answer should be comprehensive and include diverse perspectives, which are crucial for maintaining an atmosphere of neutrality, especially in public-related inquiries~\cite{habermas1991structural}.
Motivated by these considerations, we introduce the \textbf{\dataset{}} dataset, which comprises 2,941 manually annotated debatable questions.
To rigorously evaluate the aforementioned properties of model answers, each question is paired with several \textbf{partial answers}, each reflecting a single viewpoint on the debatable issue. These partial answers are generated through a three-stage pipeline and then annotated by three annotators. 
Based on the partial answers in \dataset{}, we propose the metric of \textbf{Perspective Diversity (P.D.)}, which assesses the comprehensiveness in terms of grasping multiple points-of-view in the model answer.  
Using \dataset{} along with P.D., one can reliably and efficiently evaluate the comprehensiveness of responses to debatable questions.
We also propose the metric of \textbf{Dispute Awareness (D.A.)}, which targets to identify whether the model acknowledges the debatable nature of the question in its response.
Importantly, we show that the P.D. and D.A. metrics align closely with human judgments and are fairly stable across backbone evaluator models, validating their effectiveness.

We apply \dataset{} to evaluate 12 popular LLMs, as well as retrieval-augmented generation (RAG) approaches. 
We observe while LLMs generally excel at identifying the existence of debate, their performance varies in providing comprehensive answers with diverse perspectives. 
Notably, top-notch open-source LLMs rival or even surpass some leading commercial models. 
Besides, RAG methods, though not uniformly beneficial to this task, improve closed-source model performance, likely due to better contextual leverage. 
Additionally, optimizing sampling hyperparameters and leveraging task-specific prompts can further boost performance. 
On the whole, our findings underscore the need for further refining LLMs to better interact with debatable knowledge. 

%% file: Tabs/dataset_sample.tex
\begin{table*}[ht]
\fontsize{7.75}{7.75}\selectfont
\setlength{\tabcolsep}{2pt} 
\centering
\begin{threeparttable}
\begin{tabularx}{\linewidth}{lX}
\toprule
\textbf{Field} &\textbf{Content}\\
\midrule
Question &Does birth order influence personality traits? \\
\midrule
\multirow{2}{*}{Partial Answer 1} 
&\textbf{\pov{}} Birth order does not have a meaningful and lasting effect on broad Big Five personality traits.\\ 
\cmidrule(lr){2-2}
&\textbf{\explan{}} The influence of birth order on personality traits has been a topic of interest for over a century. However, based on extensive research combining large datasets from three national panels in the United States, Great Britain, and Germany, it is evident that birth order does not have a meaningful effect on broad Big Five personality traits \ldots{} \\
\midrule
\multirow{2}{*}{Partial Answer 2} 
&\textbf{\pov{}} Firstborns score higher on intelligence and intellect.\\ 
\cmidrule(lr){2-2}
&\textbf{\explan{}} Yes, birth order does influence personality traits, particularly in the domain of intelligence and intellect. Research has consistently shown that firstborns tend to score higher on objectively measured intelligence tests 
\ldots{} \\
\midrule
\multirow{2}{*}{Partial Answer 3} 
&\textbf{\pov{}} No birth-order effects on extraversion, emotional stability, agreeableness, or conscientiousness.\\ 
\cmidrule(lr){2-2}
&\textbf{\explan{}} The influence of birth order on personality traits such as extraversion, emotional stability, agreeableness, and conscientiousness has been a topic of interest for over a century. However, recent comprehensive studies have provided substantial evidence that birth order does not significantly impact these personality traits \ldots{} \\
\bottomrule
\end{tabularx}
\end{threeparttable}
\caption{\label{tab:dataset-sample}
An example from \dataset{}, details of the explanation fields are truncated for space issues.
}
\vspace{-1em}
\end{table*}

%% file: Secs/related_work.tex
\section{Related Work}
\label{sec: related_work}

\subsection{QA with Non-fixed Answers}

Many efforts focus on QA for which there is no single fixed answer.
\textsc{AmbigQA}~\citep{min2020ambigqa} addresses ambiguous question answering by rewriting questions and generating multiple plausible answers. \textsc{SubjQA}~\citep{bjerva2020subjqa} focuses on identifying subjectivity in questions and answers within customer reviews. 
DisentQA~\citep{neeman2023disentqa} proposes to provide disentangled answers to questions where the provided context contradicts the model's knowledge.
\dataset{} differs logically from these efforts because the space of plausible answers can not be narrowed by rewriting or restricting the questions. To the best of our knowledge, DELPHI~\citep{DBLP:conf/emnlp/SunAKKXW23} is the first study on QA for debatable issues. However, DELPHI has limitations: (1) it does not provide answers for evaluation, and (2) it shallowly evaluates model performance using exploratory metrics. 
Our work represents a step forward from DELPHI by offering a comprehensive evaluation solution. We expand and refine the dataset and introduce more meaningful metrics.

\subsection{NLP on Debatable Issues}

Beyond QA, multiple lines of NLP research investigate debatable issues. One notable effort is the AI Debater, beginning with~\citet{slonim2021autonomous}'s ``IBM's Project Debater''~\citep{bar2019surrogacy}, the first AI system designed to engage humans in meaningful debates.
Another line of research focuses on controversy detection. Researchers have identified controversy in news~\citep{choi2010identifying}, online forums~\citep{chen2023anger}, and other media by analyzing sentiments~\citep{choi2010identifying, chen2023anger} or sociological features like upvote percentages~\citep{hessel2019something}.
Moreover, \citet{wan2024evidence} investigate LLMs' preference for conflicting evidence when facing controversial issues.
We distinguish ourselves from this body of research by primarily focusing on handling debatable issues in the field of QA, specifically targeting the evaluation of chatbot-like NLP systems.

\subsection{Long-form QA Evaluation}

Evaluating debatable QA falls into the topic of long-form QA evaluation. Long-form text evaluation can be categorized into two main approaches: reference-based and reference-free evaluation~\citep{xu2023critical}.
Reference-based methods require gold answers and evaluate the generated text by assessing its similarity with the gold answers~\cite{fan2019eli5, NEURIPS2021_e4d2b6e6, DBLP:conf/acl/ChiangL23}. Conversely, reference-free evaluation eliminates the necessity for a gold standard. Some studies assess the coherence and relevance of the generation concerning specified questions~\cite{DBLP:conf/naacl/FabbriWLX22, krishna-etal-2022-rankgen, xu2023critical}. Some studies use a QA-based approach to assess the quality of the generation~\cite{DBLP:journals/corr/abs-2401-15042}. In particular, there is another line of literature focusing on examining the veracity of long-form generation by utilizing external knowledge bases~\cite{stelmakh-etal-2022-asqa, min-etal-2023-factscore, wei2024long}. However, none of the aforementioned studies address the evaluation of debatable questions.

%% file: Secs/dataset.tex
\section{Curating the \dataset{} Dataset}
\label{sec: dataset}

\noindent\textbf{Dataset overview.}
\dataset{} is designed to help assessing \emph{language models' answers to debatable questions}. It contains 2,941 debatable questions, each paired with a list of \textbf{partial answers} to assist in evaluating model responses. Each partial answer addresses the question from a distinct perspective and consists of two parts: a short \emph{point-of-view} statement (\pov) and a long-form \emph{explanation} (\explan) that fully expands the \pov. An entry in \dataset{} is shown in \autoref{tab:dataset-sample}.

\begin{figure*}[ht]
    \centering
    \includegraphics[width=\linewidth]{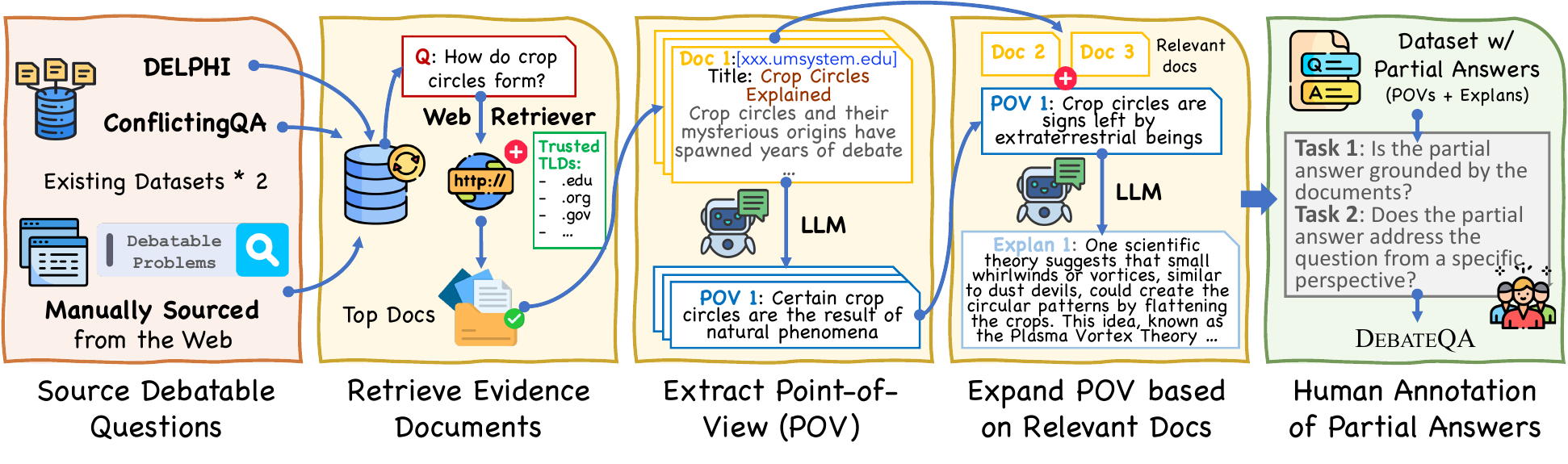}
    \vspace{-2em}
    \caption{Pipeline for curating \dataset{}. The three main components of the pipeline are highlighted in different colors: \textcolor{sourcequestions}{sourcing debatable questions}, \textcolor{collectingpartialanswer}{collecting partial answers}, and \textcolor{humanannotation}{human annotation}. Primary sources or tools used at each step are highlighted in \textbf{bold}.}
    \label{fig:dataset-pipeline}
    \vspace{-1em}
\end{figure*}

The overall procedure for curating \dataset{} is depicted in \autoref{fig:dataset-pipeline}. We first source debatable questions. Then, we apply a three-stage semi-automated pipeline to collect partial answers. Lastly, we conduct human annotation on the collected partial answers to finalize the dataset. Quality examinations happen after each step. The following sections will detail these steps.

\subsection{Sourcing Debatable Questions}

We collect debatable questions from three distinct sources. First, we repurpose two existing datasets: we select 2,281 annotated controversial questions from DELPHI~\cite{DBLP:conf/emnlp/SunAKKXW23}
and a full set of 434 questions from \textsc{ConflictingQA}~\cite{wan2024evidence}. To enrich the existing data, we further manually sourced 1,758 additional debatable questions from the Web (see~\autoref{tab:dataset-sources} for detailed sources). We then run a deduplication algorithm (see~\autoref{subsec: sources_and_deduplication} for details) to remove any duplicate questions, resulting in 3,216 questions. The final composition of sourced questions is shown in~\autoref{tab:dataset-distribution-sources}.

\subsection{Collecting Partial Answers}
\label{subsec: collect-partial-answers}

The core novelty of \dataset{} lies in evaluating models by comparing the response with multiple partial answers, rather than a single gold reference. 
One partial answer aims to answer the question from a single perspective.
This method reflects the multifaceted essence of debatable knowledge, advocating for answers that integrate diverse viewpoints~\cite{habermas1991structural, wansink2023navigating}. 
To this end, we employ a three-stage pipeline for collecting them: 
first, we collect evidence documents from trustworthy websites; 
second, we extract POVs 
from the evidence \wrt{} the question; 
finally, we expand the POVs into long-form explanations based on related evidence. The last two steps are conducted with the assistance of LLMs\footnote{We select OpenAI GPT-4~\cite{achiam2023gpt} to assist in collecting partial answers (the \texttt{gpt-4-turbo} variant).}. Together, the \emph{POVs and explanations} comprise what we call partial answers.

\subsubsection{Retrieving Trustworthy Documents}
We collect partial answers by leveraging online resources and extracting evidence from relevant web pages. However, the nature of debatable issues necessitates careful processing of these documents, as the Web can contain unveracious content.
To ensure the reliability of our partial answers, we source documents from authoritative top-level domains (TLDs), as listed in~\autoref{tab:TLDs}. This treatment helps in maintaining the reliability of the sources. We discard questions that have fewer than three documents, resulting in 2,982 questions, each supported by 3-5 of the most relevant documents. 
See~\autoref{subsec: retrieve-documents-appen} for detailed measures.

\noindent\textbf{Quality examination.} 
To assess the quality of retrieved documents, we analyze the relevancy between questions and corresponding documents. We calculate the cosine similarity between document chunks and questions. As depicted in \autoref{fig:3a-document-similarity}, the average cosine similarity for document trunks is 0.56 and there are no significant outliers, indicating high relevance and minimal noise in the documents, confirming their overall quality for serving as the basis for upcoming steps.

\subsubsection{Extracting Points-of-View}
The second stage involves extracting diverse POVs from the retrieved evidence documents. 
A POV is a concise statement that reflects the core perspective in addressing the question. 
We leverage GPT-4 to tackle this task, by applying the prompt $p_{\text{POV}}$ described in~\autoref{tab:core-prompts}, which takes the question and the concatenated documents and returns a list of diverse POVs along with the corresponding document indexes where each specific POV is originated.
The document indexes for each POV are later used for expanding the POV.
To avoid exceeding the 128K context window limit of GPT-4, we preprocess the documents by removing meaningless segments and truncating them to 120K tokens if they exceed this length. 

\noindent\textbf{Quality examination.} We verify the quality of the collected POVs on comprehensiveness \wrt{} the documents and distinctiveness among themselves. For comprehensiveness, we ensure all valid perspectives from retrieved documents are captured, with 90.4\% coverage verified manually. Distinctiveness is assured by removing duplicated POVs manually. For more details, refer to \autoref{subsec: generating-partial-answers-appen}.

\subsubsection{Expanding POV to Explanations}

The last stage involves expanding the extracted POVs into long-form explanations.
Each explanation should stand as an independent answer, elaborating on the POV and addressing the question from that perspective.
This expansion must be \emph{anchored} to the relevant information presented in the evidence documents pertaining to the specific POV being developed. 
We again leverage GPT-4 on this task, utilizing the prompt $p_{\text{Explan}}$ described in~\autoref{tab:core-prompts}. 
This prompt takes three inputs: the question, the target POV to be expanded, and the related documents obtained in the previous stage. The LLM is required to leverage only the information contained within these relevant documents to generate the explanation, minimizing the risk of hallucinations~\cite{DBLP:journals/corr/abs-2309-01219}. 
We repeat this step for all the POVs we have collected.
The pseudocode of the pipeline for collecting partial answers is deferred to \autoref{alg:partial-answers}.

\subsection{Human Annotation}
\label{subsec: annotation-dataset}

We verify the fidelity of the LLM-generated partial answers through human annotation.

\noindent \textbf{Annotation tasks.}
To thoroughly evaluate the quality of \dataset{}, we design two tasks:
\begin{packeditemize}
    \item \emph{Task 1}: Ensure that the POV and the explanation generated by the LLM are grounded in the documents. This task focuses on that the generated explanations are accurately derived from trustworthy evidence.
    \item \emph{Task 2}: Ensure that the partial answer can address the question from a certain perspective. This task assesses the utility and relevance of the partial answers, ensuring that they address the question effectively.
\end{packeditemize}

\noindent \textbf{Results and the final dataset.}
We recruit three annotators and annotate the full dataset. Inter-annotator agreement (IAA) is measured using Fleiss' Kappa~\cite{fleiss1981measurement}, yielding scores of $\kappa=0.66$ and $\kappa=0.60$ for the two annotation tasks, all indicating substantial agreement. We remove 767 partial answers deemed substandard by two or more annotators.
This suggests that GPT-4 generates faithful partial answers with a $93.4\%$ accuracy.
See~\autoref{subsec: human-annotation-appen} for details.
We employ BERTopic~\cite{DBLP:journals/corr/abs-2203-05794} to model the domain distribution of \dataset{}. The result is shown in~\autoref{fig:dataset-distribution-domain}.
To reduce computational costs for upcoming evaluation, we split \dataset{} into two splits: the \texttt{test} split with 1,000 randomly sampled questions and the \texttt{dev} set containing the remaining instances.

%% file: Secs/criteria.tex
\section{Evaluation Criteria and Metrics}
\label{sec: criteria}

\noindent \textbf{Evaluation criteria.}
For debatable questions, the quest for the \textbf{best} answer is fraught with complexity, as there exist no canonical standards.
The expectations vary: many seek a \emph{helpful} assistant that delivers credible information with no reservation, and some may pose questions merely for self-affirmation, not to say some model providers prefer a \emph{``safe''} agent to circumvent controversy.
In \dataset{}, we aim to balance helpfulness and harmlessness, with the goal of fostering open dialogues.
After reviewing interdisciplinary literature, we identify criteria that are nearly universally accepted and distill two key properties of what constitutes a \textbf{good} answer:
\begin{packeditemize}
    \item \textbf{Perspective diversity} (\emph{helpful}): how well does the answer deliver informative and credible information from diverse perspectives?
    \item \textbf{Dispute awareness} (\emph{harmless}): whether the answer recognizes the existence of debate.
\end{packeditemize}

\noindent \textbf{Evaluation metrics.}
Although the above two criteria resonate with those utilized in DELPHI~\cite{DBLP:conf/emnlp/SunAKKXW23}, we distinguish ourselves by formalizing these criteria into more sophisticated quantifiable metrics. Our methodology excels by integrating the partial answer feature of \dataset{} and outperforms DELPHI's approach by a huge margin, which will be later elaborated in \autoref{sec:PD} and \autoref{sec:DA}.
Please note that while the following two metrics both employ backbone LLMs, they may \emph{differ}; for brevity, we refer to both as \llmeval{}.

\noindent \emph{I: \criteriaone{}}.
To evaluate the model's answers against legitimate partial responses, we assess it using a smaller open-source LLM by generation, ensuring efficiency and cost-effectiveness. We apply the following metric:
\begin{align}
\fontsize{10pt}{10pt}\selectfont
\begin{split}
    \text{P.D.}&\coloneq \sum_{i=1}^n \text{PPL}(\text{PA}^i|\texttt{chatTemplate}(\text{Concat}(A,\\&\text{``Please restate.''}))),
\end{split}
\label{eq: PD}
\end{align}
\normalsize
where $ \text{PPL}(Y|X)$ is the conditional perplexity:
\fontsize{10pt}{10pt}\selectfont
\begin{equation}
    \text{PPL}(Y|X)=\exp\left(-\frac{1}{N} \sum_{i=1}^{N} \log P(y_i|X, y_{<i})\right).
\end{equation}
\normalsize
In \autoref{eq: PD}, $\text{PA}^i = \texttt{concat}(\text{POV}^i, \text{Explan}^i)$ denotes the $i^{\text{th}}$ partial answer and $A$ denotes the model's answer to the debatable question. $\texttt{chatTemplate}()$ is a chat template for prompting instruction-tuned LLMs.
Simply put, P.D. represents the aggregate perplexity of generating partial answers from a model answer. Being derived from PPL, a \textbf{lower} P.D. signifies better quality, suggesting that the model answer contains larger shares of the partial answer's content.

\noindent \emph{II: \criteriatwo{}}. 
To ascertain if the model's answer indicates that the addressed question is debatable, we craft a prompt $p_{\text{D.A.}}$, as shown in~\autoref{tab:DA-prompts}, and use it to prompt an instruction-tuned LLM. This metric is \emph{binary}, indicating awareness or lack thereof.
\begin{equation}
\fontsize{10pt}{10pt}\selectfont
  \text{D.A.} \coloneq 
  \begin{cases} 
   1 & \text{if } \mathcal{M}_{\text{eval}}(p_{\text{D.A.}}(q, Ans)) \text{ returns ``1''}, \\
   0 & \text{otherwise}.
  \end{cases}
\end{equation}
\normalsize

%% file: Secs/metric_PD.tex
\section{Empirical Examination of P.D.}
\label{sec:PD}

To assess the reliability of P.D., we empirically examine its alignment with human preferences and stability \wrt{} backbone models and prompts.

\subsection{P.D. Well-Aligns Human Preferences}
\label{subsec:PD-alignment}

\noindent \textbf{Collecting human preferences.}
We collect 500 model answers from five LLMs to 100 randomly sampled questions in \dataset{}-\texttt{test}, then have three rank the answers pairwise. 
\begin{wrapfigure}{r}{0.45\linewidth}
    \centering
    \includegraphics[width=\linewidth]{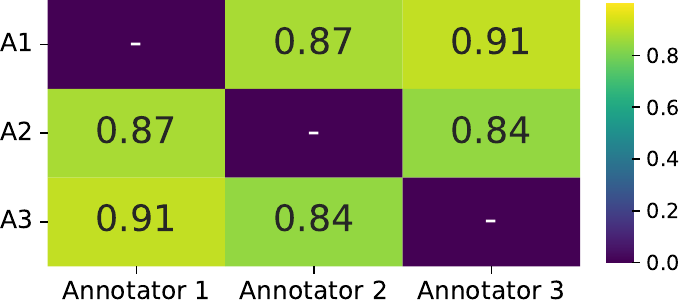}
    \vspace{-2em}
    \caption{Correlation among annotators assessed by Spearman's $\rho$.}
    \label{fig:correlation-preference}
    \vspace{-1em}
\end{wrapfigure}
Using pairwise preferences and Spearman's $\rho$ correlation~\cite{Zar_2005}, we find strong agreement among three annotators with $\rho >0.8$, as shown in \autoref{fig:correlation-preference}. This consensus allows us to assess the alignment of evaluator judgments with human preferences effectively. More details are described in \autoref{subsubsec:annotation-preferences}.

\noindent \textbf{Compared baselines.}
We compare P.D. with notable text evaluation baselines. Many of these require a language model as backbones:
\begin{packeditemize}
    \item \textbf{\emph{Prompt-based metrics}}: These baselines always require an advanced instruction-tuned LLM, we consider three of them: (1) Direct-Score: using an LLM to assign a Likert scale score (1-5) using the \emph{same} instruction we present to human annotators; (2) G-Eval~\cite{DBLP:conf/emnlp/LiuIXWXZ23}: a prompt-based evaluation framework that assesses the quality of generated texts by incorporating chain-of-thoughts (CoT)~\cite{DBLP:conf/nips/Wei0SBIXCLZ22} and a form-filling paradigm; (3) Num-of-POVs: using an LLM to count the number of perspectives in the model answers. This metric can be considered an improved metric over the ``Comprehensiveness Answer Rate'' metric introduced in DELPHI~\cite{DBLP:conf/emnlp/SunAKKXW23}.
    \item \textbf{\emph{Similarity-based metrics}}: We pick two traditional statistical metrics, BLEU~\cite{DBLP:conf/acl/PapineniRWZ02} and ROUGE-L~\cite{lin2004rouge} and two neural metrics, BARTScore~\cite{NEURIPS2021_e4d2b6e6} and SemScore~\cite{DBLP:journals/corr/abs-2401-17072}.
\end{packeditemize}
Refer to \autoref{subsubsec:baseline-evaluation} for details of these baselines.

\noindent\textbf{Results.}
We apply {ELO ratings}
to establish a human preference ranking and then measure the correlation with metrics' rankings using Spearman's $\rho$ and Kendall's $\tau$~\cite{KENDALL}.
The average results of the annotators' individual correlation are reported in~\autoref{tab: alignment-experiment-result}.
We observe that P.D. significantly outperforms DELPHI's metric and G-Eval powered by 
GPT-4o. 
Using the same small LLM (Phi-3 mini 128k) as the backbone, P.D. outperforms G-Eval by a huge margin,
underscoring P.D.'s exceptional alignment with human judgment. 
Overall, P.D. is effective and economical.
We further provide case studies in \autoref{subsubsec:PD-casestudy}.

\begin{table}[ht]
\fontsize{9}{8}\selectfont
\setlength{\tabcolsep}{2pt} 
\centering
\vspace{-1em}
\begin{threeparttable}
\begin{tabular}{lccc} 
\toprule
\textbf{Metric} &Backbone Model &$\rho$ &$\tau$ \\
\midrule
\multicolumn{4}{c}{\emph{Prompt-based metrics}}\\
\midrule
Direct-Score & GPT-4o & 0.692 & 0.671 \\
G-Eval & Phi-3 mini 128k &-0.003 & -0.028 \\
G-Eval & GPT-3.5 Turbo &0.593 & 0.531 \\
G-Eval & GPT-4o & 0.706  & 0.634  \\
Num-of-POVs & GPT-4o & 0.398 & 0.345 \\
\midrule
\multicolumn{4}{c}{\emph{Similarity-based metrics}}\\
\midrule
BLEU & - & 0.667 & 0.565  \\
ROUGE-L & - & 0.537 & 0.451 \\
BARTScore & BART Large~(\citeyear{DBLP:conf/acl/LewisLGGMLSZ20}) & 0.476 & 0.397 \\
SemScore & all-mpnet-base-v2~(\citeyear{DBLP:conf/emnlp/ReimersG19}) & 0.447 & 0.375 \\
\textbf{P.D.} &Phi-3 mini 128k &  0.733  & 0.701 \\
\textbf{P.D.} &GPT-2 (117M) &  \textbf{0.825}  & \textbf{0.748}  \\
\textbf{P.D.} &Qwen2 0.5B &  \underline{0.820}  &\underline{0.742} \\
\bottomrule
\end{tabular}
\end{threeparttable}
\caption{\label{tab: alignment-experiment-result} Alignment of various evaluation metrics with human preferences. The top-performing metric is highlighted in \textbf{bold} and the runner-up is \underline{underlined}. $\rho$: Spearman's $\rho$, $\tau$: Kendall's $\tau$.}
\vspace{-1em}
\end{table}

\subsection{P.D. is Stable \wrt{} Backbone Models}

\begin{figure}[ht]
    \centering
    \includegraphics[width=\linewidth]{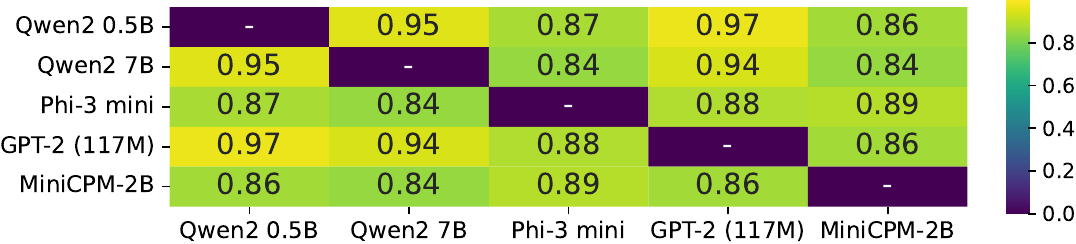}
    \vspace{-2em}
    \caption{Kendall's $\tau$ correlations of the P.D. metric using different backbone models.}
    \vspace{-1em}
    \label{fig:2b-PD-kendall-correlations-across-models}
\end{figure}

To verify the stability of P.D. \wrt{} different backbone models, we configure P.D. with five different LLMs and compute pairwise Kendall's $\tau$ among the resulting rankings, for the same set of model responses collected in \autoref{subsec:PD-alignment}. The results in~\autoref{fig:2b-PD-kendall-correlations-across-models} show that the rankings by P.D. with different backbone models are highly consistent.

\subsection{P.D. is Stable \wrt{} Prompts}
\label{subsec: PD-stability-prompts}

\begin{figure}[ht]
    \centering
    \includegraphics[width=\linewidth]{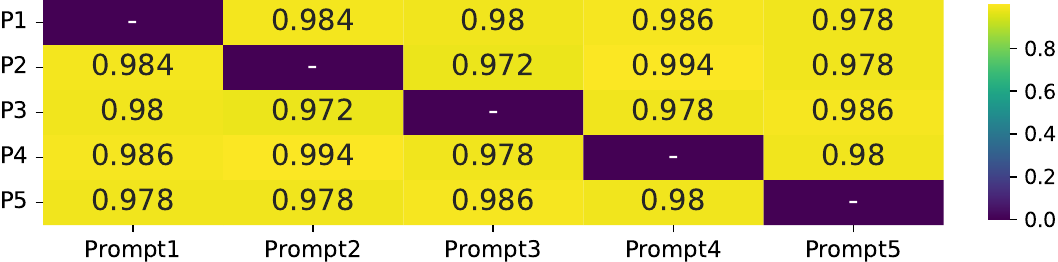}
    \vspace{-2em}
    \caption{Kendall's $\tau$ correlations of the P.D. metric using different prompts.}
    \vspace{-1em}
    \label{fig:PD-kendall-correlations-across-prompts}
\end{figure}

Remember in \autoref{eq: PD}, there is a prompt ``Please Restate'' wrapping the model answer. To verify the stability of P.D. \wrt{} different prompts, we configure P.D. with five different prompts shown in~\autoref{tab:PD-stability-prompts} and compute pairwise Kendall's $\tau$ among the resulting rankings, for the same set of model responses collected in \autoref{subsec:PD-alignment}. The results in~\autoref{fig:PD-kendall-correlations-across-prompts} show that the rankings by P.D. with different prompts have excellent consistency.

%% file: Secs/metric_DA.tex
\section{Empirical Examination of D.A.}
\label{sec:DA}

We assess D.A.'s reliability by examining its accuracy grounded by human judgments and stability \wrt{} backbone models.

\subsection{D.A. is Accurate}
\label{subsec:DA-accuracy}

To obtain the ground truth of the verdicts of the 500 responses from \autoref{subsec:PD-alignment}, three authors manually annotate them by assigning binary labels. The annotation has an inter-annotator agreement of 0.79 evaluated by Fleiss' Kappa.
\begin{table}[ht]
\centering
\setlength{\tabcolsep}{2pt} 
\fontsize{9}{8.5}\selectfont
\begin{threeparttable}
\begin{tabular}{lccccc}
\toprule
\textbf{Metric} & Backbone Model &Acc &F1 &AUROC 
&MCC 
\\
\midrule
\textbf{D.A.} & Qwen2 1.5B& 0.740& 0.845 & 0.530
& 0.093
\\
\textbf{D.A.} & MiniCPM-2B-dpo& \underline{0.770}  &0.857& \underline{0.621}&\underline{0.289}
\\
\textbf{D.A.} & Llama3 8B& \underline{0.770}  &\textbf{0.869}&0.521 & 0.179
\\
\textbf{D.A.} & Gemma 2 9B& 0.740  &0.833& 0.615& 0.246
\\
\textbf{D.A.} & Phi-3 medium& \textbf{0.800}  &\underline{0.868}& \textbf{0.726}&\textbf{0.452} 
\\
\cmidrule{1-6}
D.A.-ZS&  Phi-3 medium & 0.530& 0.434 &0.565 
& 0.146
\\
Ack. & - &0.220 & 0.049 & 0.430 & -0.252 
\\
\bottomrule
\end{tabular}
\end{threeparttable}
\caption{\label{tab:awarness-experiment} Based on human annotations as the ground truth, we assess D.A.'s accuracy. D.A.-ZS: D.A. with a zero-shot prompt $p_{\text{D.A.-ZS}}$, Ack.: the ``Acknowledge'' metric from DELPHI.}
\vspace{-1em}
\end{table}
With these manually labeled outcomes as the ground truth, we calculate the accuracy, F1 score, and AUROC \cite{hanley1982meaning} and Matthews Correlation Coefficient (MCC; \citealt{matthews1975comparison}) 
for D.A. with different backbone models, the ``Acknowledge'' metric referenced in DELPHI (refer to \autoref{subsec:DA-accuracy-appen} for details), and a simplified version of D.A. with a zero-shot prompt $p_{\text{D.A.-ZS}}$ without in-context demonstrations.
The results in \autoref{tab:awarness-experiment} demonstrate the superiority of our D.A. metric over the metric from DELPHI, and the necessity of including demonstrations in the prompt $p_{\text{D.A.}}$. Upon a case study for D.A.-ZS, we find that the Phi-3 mini 128k model occasionally deviates from our instruction by failing to use 0 or 1 for its judgment, thereby diminishing its effectiveness.

\subsection{D.A. is Stable \wrt{} Backbone Models}
\begin{figure}[ht]
    \centering
    \includegraphics[width=\linewidth]{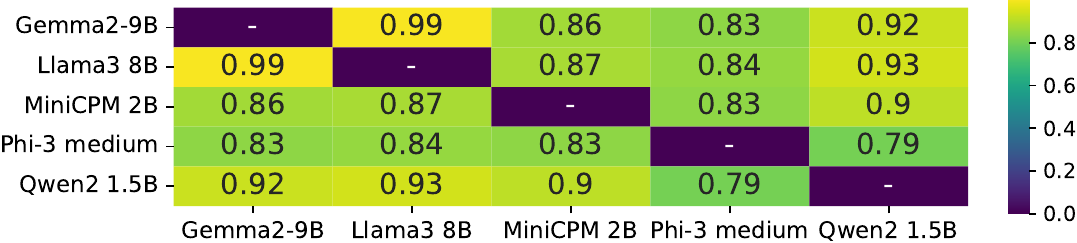}
    \vspace{-2em}
    \caption{Agreements of the D.A. metric across different backbone models.}
    \label{fig:2d-DA-agreement-of-models}
    \vspace{-1em}
\end{figure}

Given the robust design of prompt $p_{\text{D.A.}}$, ensuring D.A.'s performance, we advise utilizing the \emph{standard} prompt in \autoref{tab:DA-prompts}. Our focus here is on confirming D.A.'s stability across various backbone models.
We set D.A. with five instruction-following LLMs and analyze pairwise agreements for the decision made between each two LLMs based on the model responses collected in \autoref{subsec:PD-alignment}. The result in \autoref{fig:2d-DA-agreement-of-models} demonstrates the consistency of D.A. among different backbone models.

%% file: Secs/experiments.tex
\section{Experiments}
\label{sec: experiments}

\input{Tabs/experiment_test_models}

\subsection{Experimental Setup}
\label{subsec: experimental-setup}

We evaluate a wide range of 12 LLMs on \dataset-\texttt{test} using P.D. and D.A., including closed commercial LLMs and open-source ones. 
We also assess several RAG approaches.

\noindent \textbf{Evaluated models.} We evaluate the following LLMs: GPT-4o~\cite{openai2024gpt4o}, GPT-4o mini~\cite{openai2024gpt4omini}, GPT-3.5 Turbo~\cite{openai2023gpt35turbo}, Claude 3.5 Sonnet~\cite{anthropic2024claude35sonnet}, Llama3 Instruct 8B/70B~\cite{meta2024llama3}, Qwen2 0.5B/1.5B/7B~\cite{qwen2023qwen2}, Phi-3 mini 128k 3.8B and Phi-3 small 128k 7B~\cite{DBLP:journals/corr/abs-2404-14219}, and Gemma2 9B \cite{google2024gemma}. All models are the instruction or chat fine-tuned versions.

\noindent \textbf{Evaluators.} We select multiple language models as the backbone for our metrics. For evaluating P.D., we select Qwen2 0.5B and GPT-2 base 117M~\cite{radford2019language} as \llmeval{}.
For assessing D.A., a competent LLM with instruction-following ability is a must. We select Phi-3 medium 128k 14B and Qwen2 1.5B. We select those four models because their performance is showcased in \autoref{subsec:PD-alignment} and \autoref{subsec:DA-accuracy}, respectively.

\noindent \textbf{Generation configuration.}
In the main experiments, when testing the LLMs, we provide the questions with a minimalistic QA prompt, as shown in \autoref{tab:prompt-no-rag}, which instructs the LLMs without any hint that they are debatable. 
We believe this approach more accurately reflects the typical user interaction with chatbots. 
For all models, we configure top-$p=0$ to enable greedy decoding and stock chat templates including \llmeval{}.

\subsection{Main Results}
\label{subsec:main-results}

Main evaluation results 
can be found in 
\autoref{tab:main-results}. 
We summarize our key findings as follows:
\begin{packeditemize}
    \item \textbf{Larger models generally outperform small ones.} Generally, larger models perform well in terms of both P.D. and D.A. metrics. Large state-of-the-art LLMs such as GPT-4o and Llama3 70B demonstrate the strongest performance, while tiny to small LLMs are almost always at the bottom. Besides, for LLMs with the same architecture (\eg, Qwen 2 0.5B/1.5B/7B), larger models always outperform smaller ones.
    \item \textbf{Gaps between closed and open models are not clear.} With the exception of the superb performance of GPT-4o, we have observed that numerous open-source LLMs are outperforming other capable closed commercial LLMs. Specifically, open-source models like Llama3 70B and Gemma2 9B nearly match the performance of GPT-4o in handling debatable questions. This might indicate that the performance on handling debatable issues does not demand as stringent capabilities from the models as more difficult tasks such as reasoning.
    \item \textbf{Deficiencies in delivering comprehensive responses.} We observe significant shortcomings in weaker models' ability to furnish comprehensive answers that encompass a variety of perspectives. For instance, the worst-performing Qwen2 0.5B's answers are around $3$ to $9 \times$\footnote{These values are approximated with the P.D. values \wrt{} Two different \llmeval{}, refer to \autoref{sec:interpretation} for details.} worse than GPT-4o in terms of recovering the information in partial answers. 
    \item \textbf{Models excel in recognizing debate.} We find that even the lowest-performing model, namely Qwen2 0.5B, demonstrates a D.A. over 70\% in recognizing the contentiousness of debatable questions. This indicates that even the performance of the worst-performing model is not as poor as it might seem.
\end{packeditemize}

\begin{wrapfigure}{r}{0.45\linewidth}
    \centering
    \vspace{-1em}
    \includegraphics[width=\linewidth]{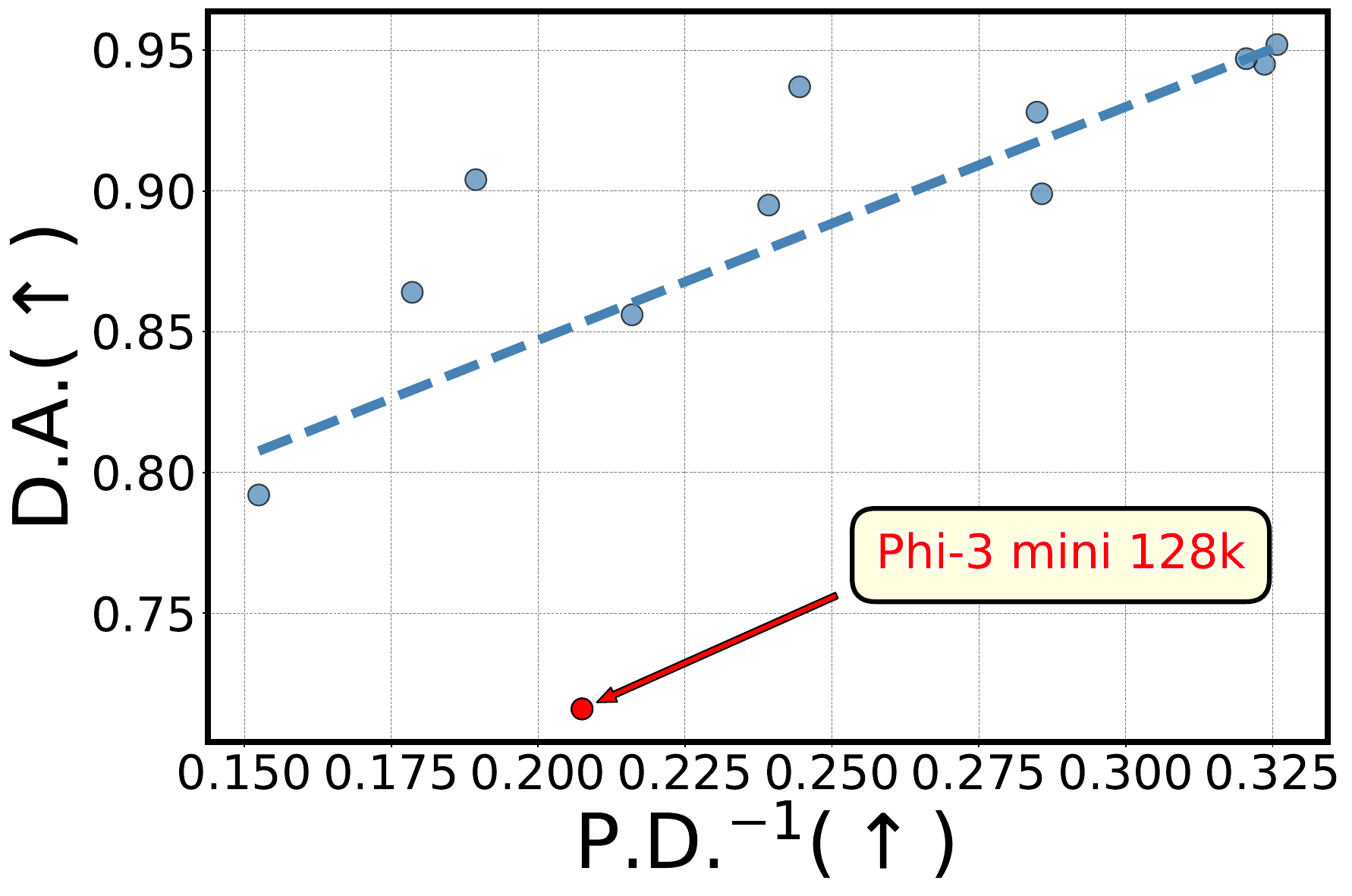}
    \vspace{-2em}
    \caption{Correlation between P.D. and D.A. for the tested LLMs.}
    \label{fig:correlation-between-PD-DA}
    \vspace{-1em}
\end{wrapfigure}

\noindent\textbf{Correlation between the P.D. and D.A. metrics.}
To investigate the correlation between P.D. and D.A. for the tested models, we plot a visualization of the results in \autoref{fig:correlation-between-PD-DA}, which aids in understanding how these two metrics relate to each other across various models.
The figure indicates a positive correlation between P.D. and D.A., suggesting that models with greater perspective diversity are also more aware of the debate in question, aligning with our expectations.

\begin{figure}
    \centering
    \vspace{-1em}
    \includegraphics[width=1\linewidth]{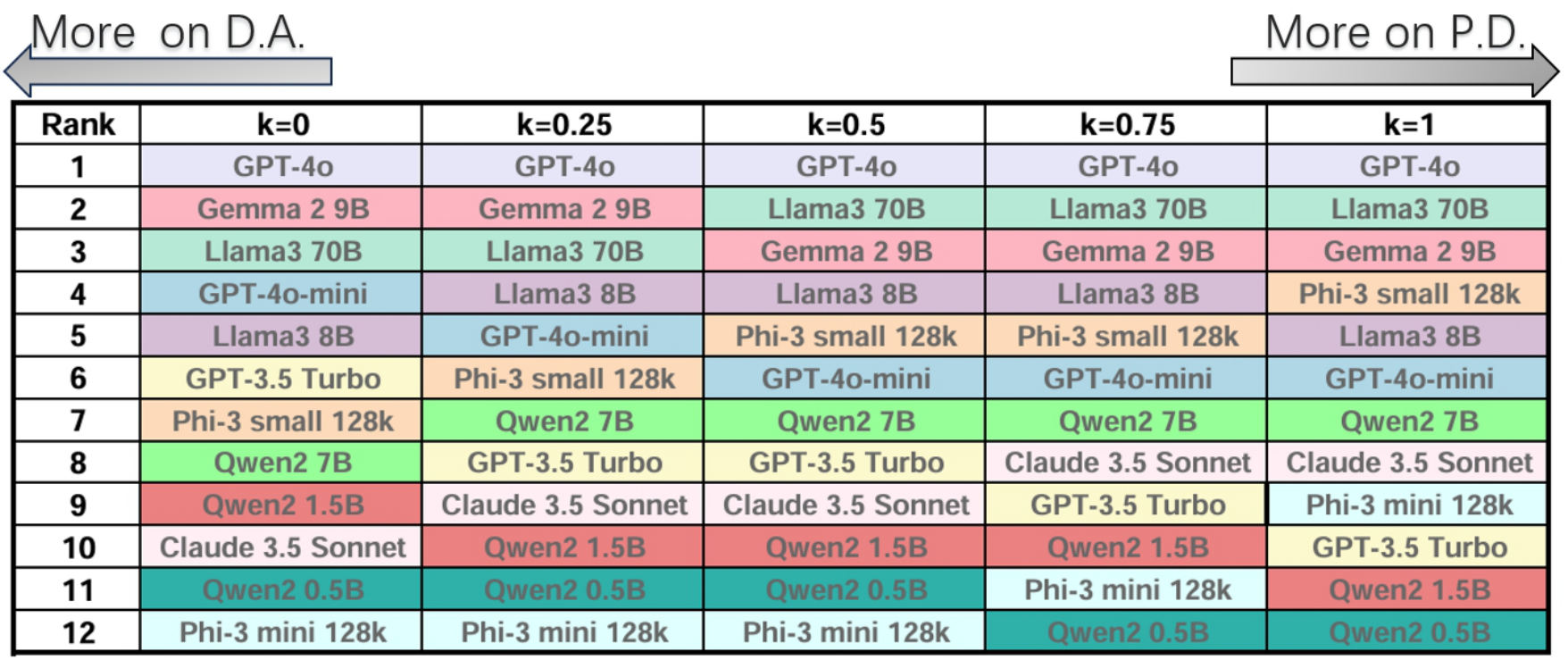}
    \vspace{-2em}
    \caption{Rankings of weighted average scores of models at different ratios. $k$ is the proportional coefficient.}
    \label{fig:1a-average-scores-for-models}
    \vspace{-1em}
\end{figure}

\noindent\textbf{Integrating the P.D. and D.A. metrics.} 
To demonstrate the models' general capability in addressing debatable questions, we integrate the two metrics with varying weights.
To ensure that the P.D. scores reflect the same performance favorability as the D.A. scores, we first take the reciprocal of the P.D. scores.
Then, we leverage z-normalization\footnote{\url{https://en.wikipedia.org/wiki/Standard_score}} to normalize both the reciprocal P.D. score and the original D.A. score.
The final weighted average score (W.A.G.) is calculated as:
\begin{equation}
\fontsize{9}{9}\selectfont
    \text{W.A.G.} = k \cdot \texttt{Z-norm}(\text{P.D.}^{-1})+ (1-k) \cdot \texttt{Z-norm}(\text{D.A.}),
\end{equation}
\normalsize
where the $k$ is the proportional coefficient and $\texttt{Z-norm}()$ refers to z-normalization. We show the ranking of the weighted average scores for the models at different $k$ in~\autoref{fig:1a-average-scores-for-models}.

\subsection{Effect of More Specific Prompts}

In our main experiments, we use a simple QA prompt ($p_{\text{basic}}$ in \autoref{tab:system-prompts}) that does not highlight the debatable nature of the questions or demand comprehensive answers. 
To evaluate the models' full potential, we further test five LLMs with more detailed prompts.
We employ three system prompts---$p_{\text{basic}}$, $p_{\text{comprehensive}}$, and $p_{\text{detailed}}$---to elicit model responses at varying levels of detail, as shown in \autoref{tab:system-prompts}. 
Using 200 randomly sampled questions from \dataset-\texttt{test}, we compare the average P.D. and D.A. scores across the five selected LLMs. 
The results for these prompts are presented in \autoref{tab:1c-PD}.
We find that even the relatively simple $p_{\text{comprehensive}}$ prompt significantly improved the performance for all five models. 
We conclude that more specific prompts, \ie, inform the model of the debatable nature and request for detailed responses, can enhance LLMs' performance in answering debatable questions. 
This finding aligns with our expectations and suggests that LLM users can benefit from well-crafted prompts when seeking answers to contentious issues from LLMs.

\begin{table}[ht]
\fontsize{9}{9}\selectfont
\setlength{\tabcolsep}{2pt} 
\centering
\begin{threeparttable}
\begin{tabular}{lcccccc}
\toprule
\multirow{2}{*}{\textbf{Model}} & \multicolumn{3}{c}{P.D. ($\downarrow$)} & \multicolumn{3}{c}{D.A. ($\uparrow$)} \\
\cmidrule(lr){2-4} \cmidrule(lr){5-7} 
 & $p_{\text{b}}$ & $p_{\text{c}}$ & $p_{\text{d}}$  & $p_{\text{b}}$ & $p_{\text{c}}$ & $p_{\text{d}}$ \\
\midrule
GPT-4o mini & 3.91 & 2.13 & \textbf{2.09} &  0.915 & 0.955 & \textbf{0.97}\\
Claude 3.5 Sonnet & 4.63 & 3.14& \textbf{2.35}& 0.865 & 0.925 & \textbf{0.98} \\
Llama3 8B & 3.42 & 2.58 & \textbf{2.51}& 0.855 & 0.935& \textbf{0.985}  \\
Qwen2 7B & 4.16 & 2.78 & \textbf{2.76}& 0.855 &0.915  & \textbf{0.965}\\
Phi-3 mini 128k &4.71 & 3.40 & \textbf{2.88} &0.765 & 0.925& \textbf{0.96}\\
\bottomrule
\end{tabular}
\end{threeparttable}
\caption{\label{tab:1c-PD} Effect of various prompts on P.D. scores and D.A. scores. $p_{\text{b}}$: $p_{\text{basic}}$, $p_{\text{c}}$: $p_{\text{comprehensive}}$, $p_{\text{d}}$: $p_{\text{detailed}}$.}
\vspace{-2em}
\end{table}

\subsection{Evaluation for RAG Approaches}

In this section, we assess the influence of RAG 
methods on performance within \dataset{}.

\subsubsection{Effect of RAG Strategy}

We examine the effects of two popular RAG strategies, \emph{vanilla RAG}~\cite{DBLP:conf/nips/LewisPPPKGKLYR020} and \emph{ReAct}~\cite{DBLP:conf/iclr/YaoZYDSN023}.
In vanilla RAG, we pick the top-10 most relevant documents from the retrieval results via Google Custom Search API\footnote{\url{https://developers.google.com/custom-search}}.
ReAct employs an agent-based approach, leveraging Claude 3.5 Sonnet to interleave reasoning with document retrieval, strategically selecting up to 9 document chunks to improve problem-solving.
Both methods utilize the prompt in \autoref{tab:RAG-prompts} to assemble the question and the retrieved trunks. 
Refer to~\autoref{subsec: rag-config-appen} for details.

We assess the performance of five LLMs by evaluating their responses to 100 randomly sampled debatable questions from \dataset{}-\texttt{test} using two distinct RAG strategies. With the results detailed in \autoref{tab:1d-rag-setup}, we conclude:
\begin{packeditemize}
    \item LLMs with RAG do not consistently improve in answering debatable questions, but closed-source models see more notable benefits, possibly due to better context utilization despite potential noise in retrieved content chunks.
    \item Among the two RAG strategies, ReAct consistently outperforms Vanilla RAG, even though it uses fewer document chunks (9 vs. 10). This advantage can be attributed to ReAct's more strategic approach to acting based on the previously retrieved information, which leads to more precise and relevant documents retrieved.
\end{packeditemize}

\begin{table}[ht]
\fontsize{9.5}{9.5}\selectfont
\setlength{\tabcolsep}{2pt} 
\centering
\begin{threeparttable}
\begin{tabular}{lccc}
\toprule
\multirow{2}{*}{\textbf{Model}} & \multicolumn{3}{c}{P.D. ($\downarrow$ \llmeval{}=Qwen2 0.5B )} \\
\cmidrule{2-4}
 & No RAG & Vanilla RAG & ReAct \\
\midrule
GPT-4o mini & 4.02 & 3.94  & \textbf{3.70}  \\
Claude 3.5 Sonnet & 4.63 & 4.12 & \textbf{3.65}  \\
Llama3 8B  &\textbf{3.55} & 4.01  &  3.99 \\
Qwen2 7B & \textbf{3.79}& 5.96 &5.29  \\
Phi-3 mini 128k & \textbf{4.82} &  7.01 & 6.86 \\
\bottomrule
\end{tabular}
\end{threeparttable}
\caption{\label{tab:1d-rag-setup} Effect of two RAG strategies on P.D. scores.}
\vspace{-1em}
\end{table}

\subsubsection{Effect of RAG Source Documents}

Considering that the performance of RAG is highly dependent on the quality of the retrieved documents, we explore whether restricting RAG to utilize trustworthy documents would yield better results. We retrieve only on web pages under trustworthy TLDs listed in \autoref{tab:TLDs}.
The results in~\autoref{tab:1e-PD} demonstrate that RAG on trustworthy sources leads to better results.
This highlights the significance of source quality in RAG for debatable QA, emphasizing that utilizing trustworthy documents improves LLM response quality in responding to sensitive topics.

\begin{table}[ht]
\fontsize{9.5}{9.5}\selectfont
\setlength{\tabcolsep}{3pt} 
\centering
\begin{threeparttable}
\begin{tabular}{lccc}
\toprule
\multirow{3}{*}{\textbf{Model}} & \multicolumn{2}{c}{P.D. ($\downarrow$ \llmeval{}=Qwen2 0.5B )} \\
\cmidrule{2-3}
 & Vanilla RAG & RAG w. T. Docs \\
\midrule
GPT-4o mini & 3.77  & \textbf{3.63}  \\
Claude 3.5 Sonnet  & 3.92 & \textbf{3.54}  \\
Llama3 8B  & 3.78  & \textbf{3.62} \\
Qwen2 7B & 5.91 & \textbf{5.57}  \\
Phi-3 mini 128k  &  6.77 & \textbf{6.50} \\
\bottomrule
\end{tabular}
\end{threeparttable}
\caption{\label{tab:1e-PD} Effect of RAG sources on P.D. scores. RAG w. T. Docs: RAG using trustworthy documents.}
\vspace{-1em}
\end{table}

\subsection{Effect of Decoding Hyperparameters}
\label{subsec: decoding}

\begin{figure}[htbp]
    \centering
        \vspace{-2em}
    \subfloat[GPT-4o-mini]{
        \includegraphics[width=0.4\linewidth]{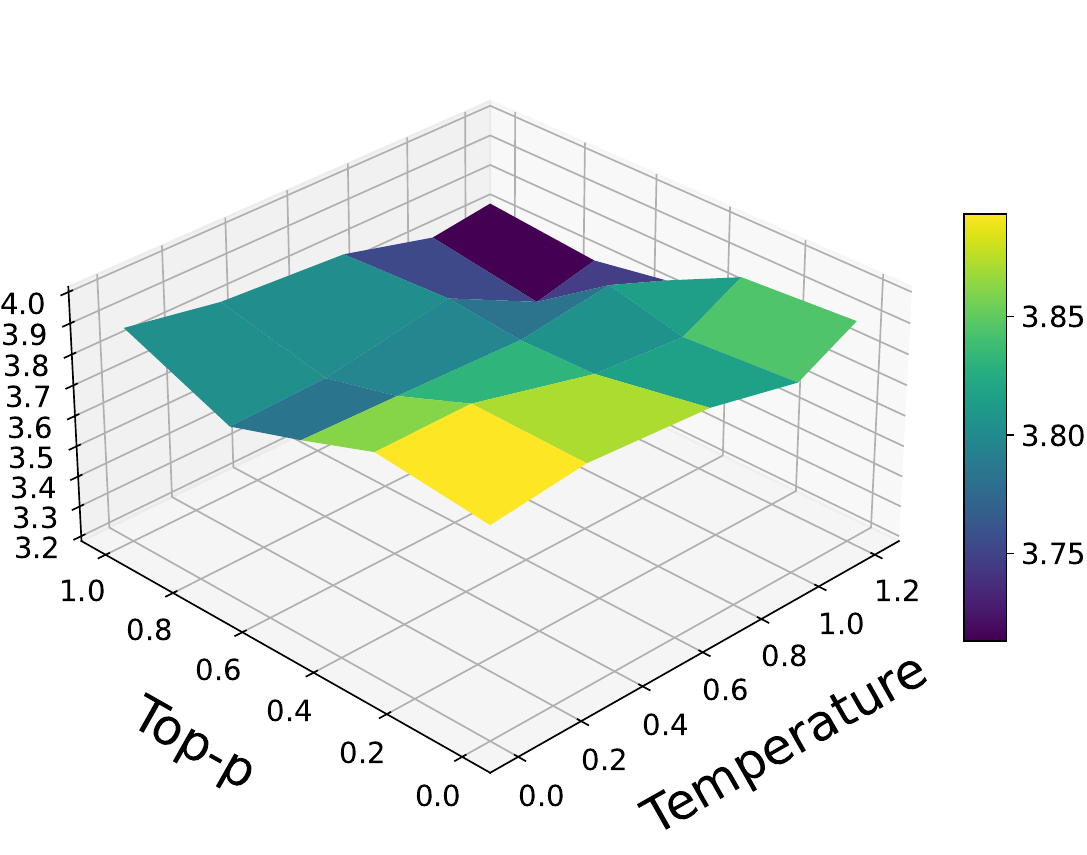}
        \label{fig:1b-4omini-PD}
    }
        \hspace{0mm}
    \subfloat[Llama3 8B]{
        \includegraphics[width=0.4\linewidth]{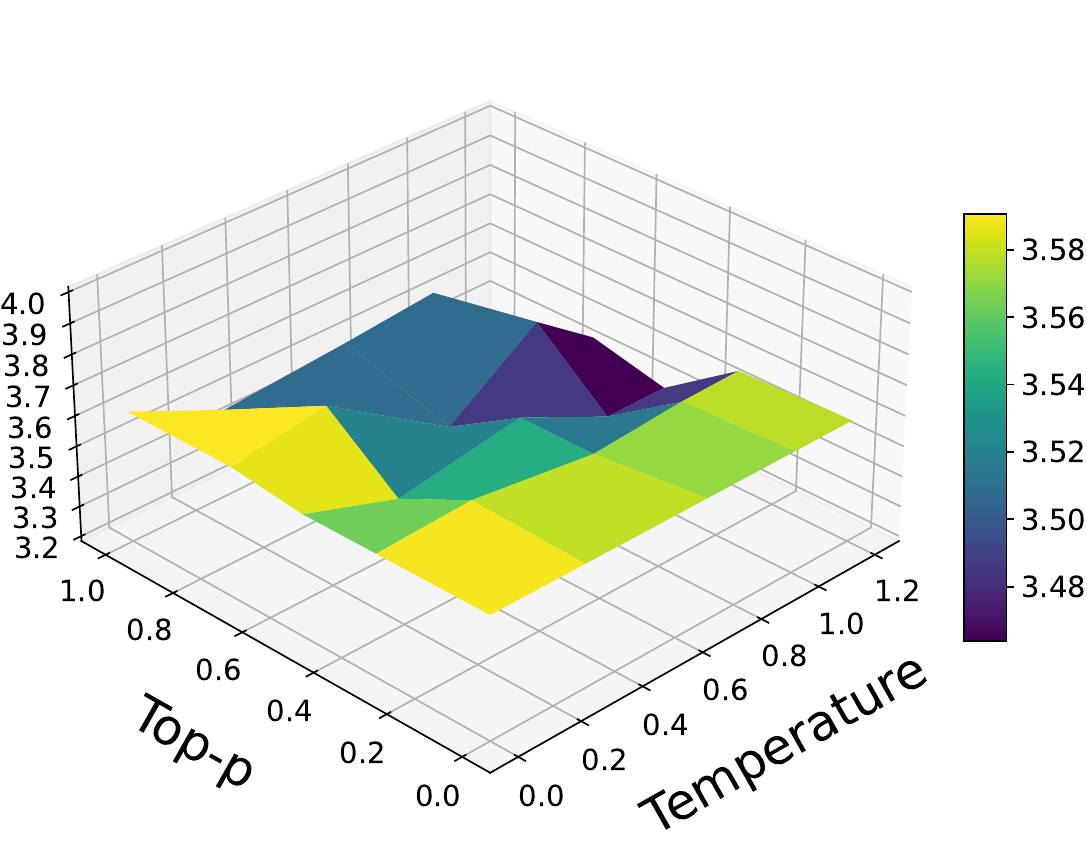}
        \label{fig:1b-llama3-PD}
    }
    \vspace{-1em}
    \caption{Average P.D. score ($\downarrow$) for answers from different LLMs with corresponding hyperparameters.}
    \label{fig:1b-PD}
\end{figure}

\begin{figure}[htbp]
    \centering
    \vspace{-2em}
    \subfloat[GPT-4o-mini]{
        \includegraphics[width=0.4\linewidth]{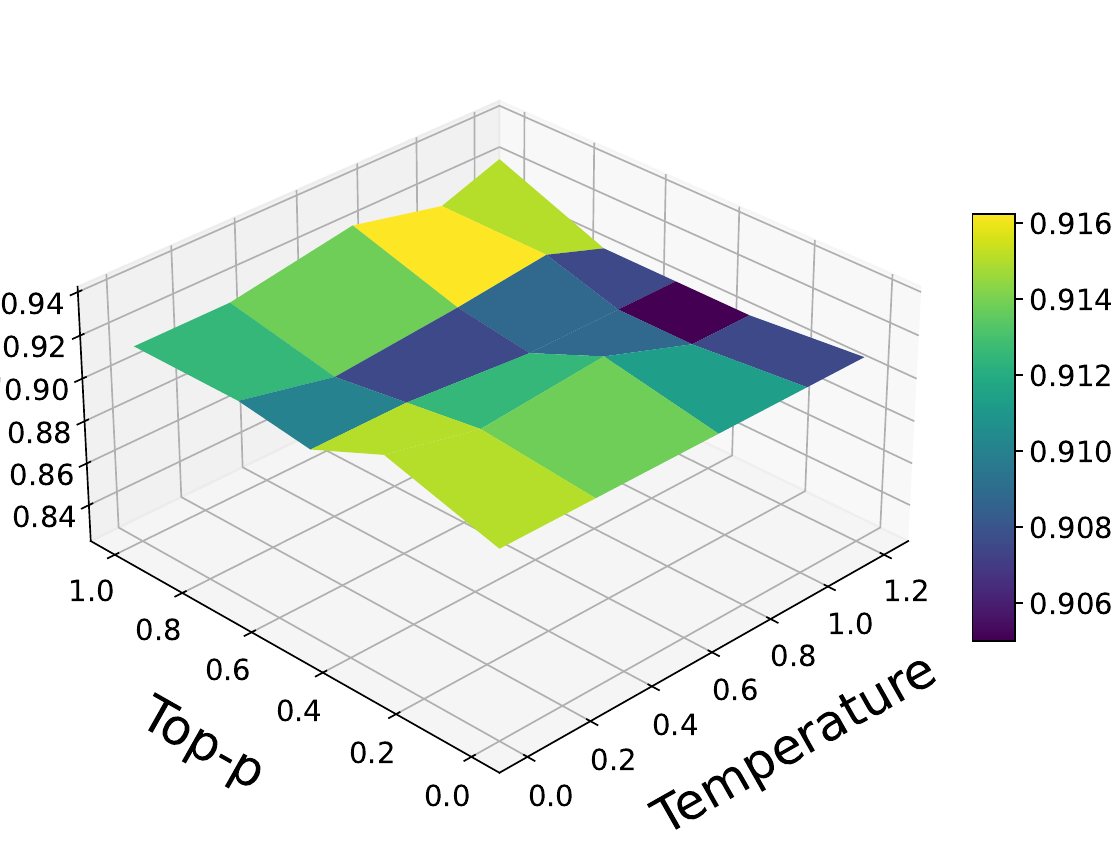}
        \label{fig:hyper1}
    }
    \hspace{0mm}
    \subfloat[Llama3 8B]{
        \includegraphics[width=0.4\linewidth]{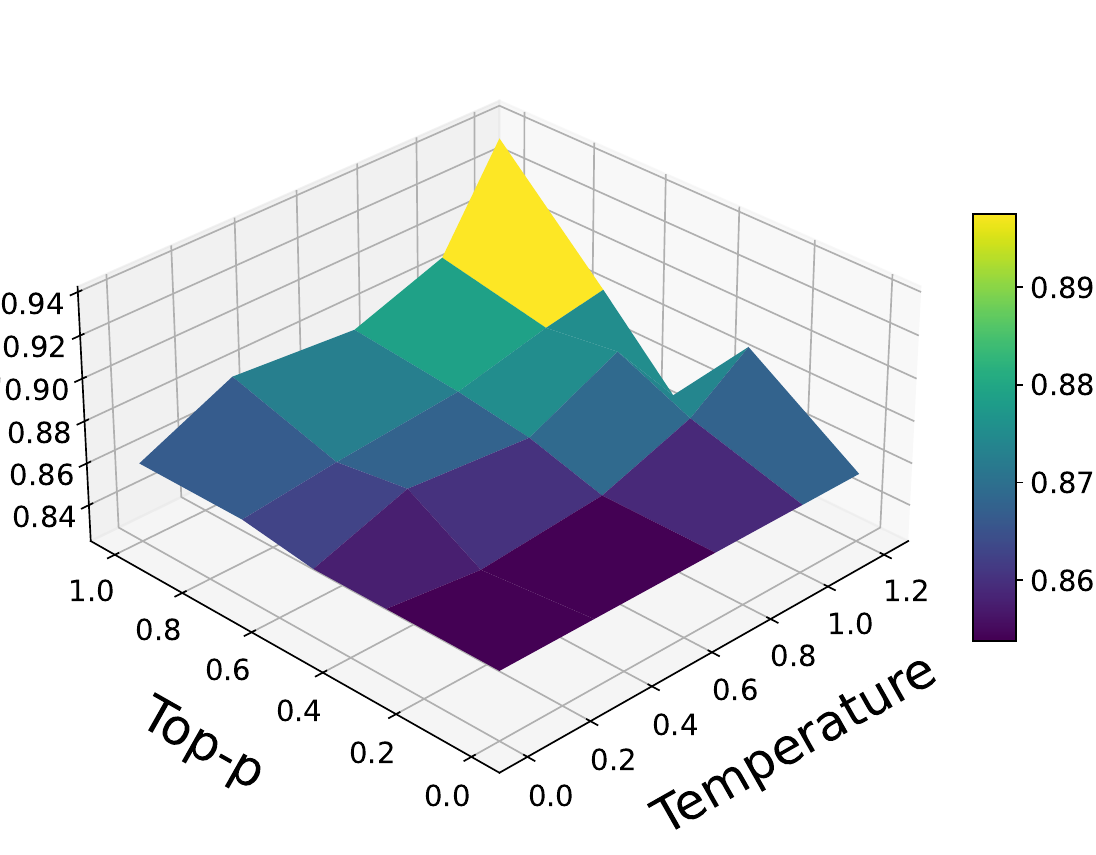}
        \label{fig:hyper2}
    }
    \vspace{-1em}
    \caption{Average D.A. score ($\uparrow$) for answers from different LLMs with corresponding hyperparameters.}
    \label{fig:1b-DA}
    \vspace{-1em}
\end{figure}

In the main experiments, we configure all LLMs to use greedy decoding, which, while straightforward, can restrict the diversity and creativity of LLM outputs~\cite{DBLP:conf/iclr/HoltzmanBDFC20}. 
To assess the impact of various decoding hyperparameters using sampling decoding on models' performance, we select a range of five different temperatures and top-$p$ values. 
The results of P.D. and D.A. scores are presented in \autoref{fig:1b-PD} and \autoref{fig:1b-DA}, respectively.
The plots indicate that higher temperature and top-$p$ values generally prompt LLMs to produce more well-rounded responses to debatable questions, enhancing performance on both metrics. This suggests that sampling configurations that allow for a broader selection of lower-probability tokens can lead to improved outcomes.


\subsection{Effect of the Length of Generation}

\begin{figure}[t]
    \centering
    \subfloat[P.D. Scores ($\downarrow$)]{
        \includegraphics[width=0.825\linewidth]{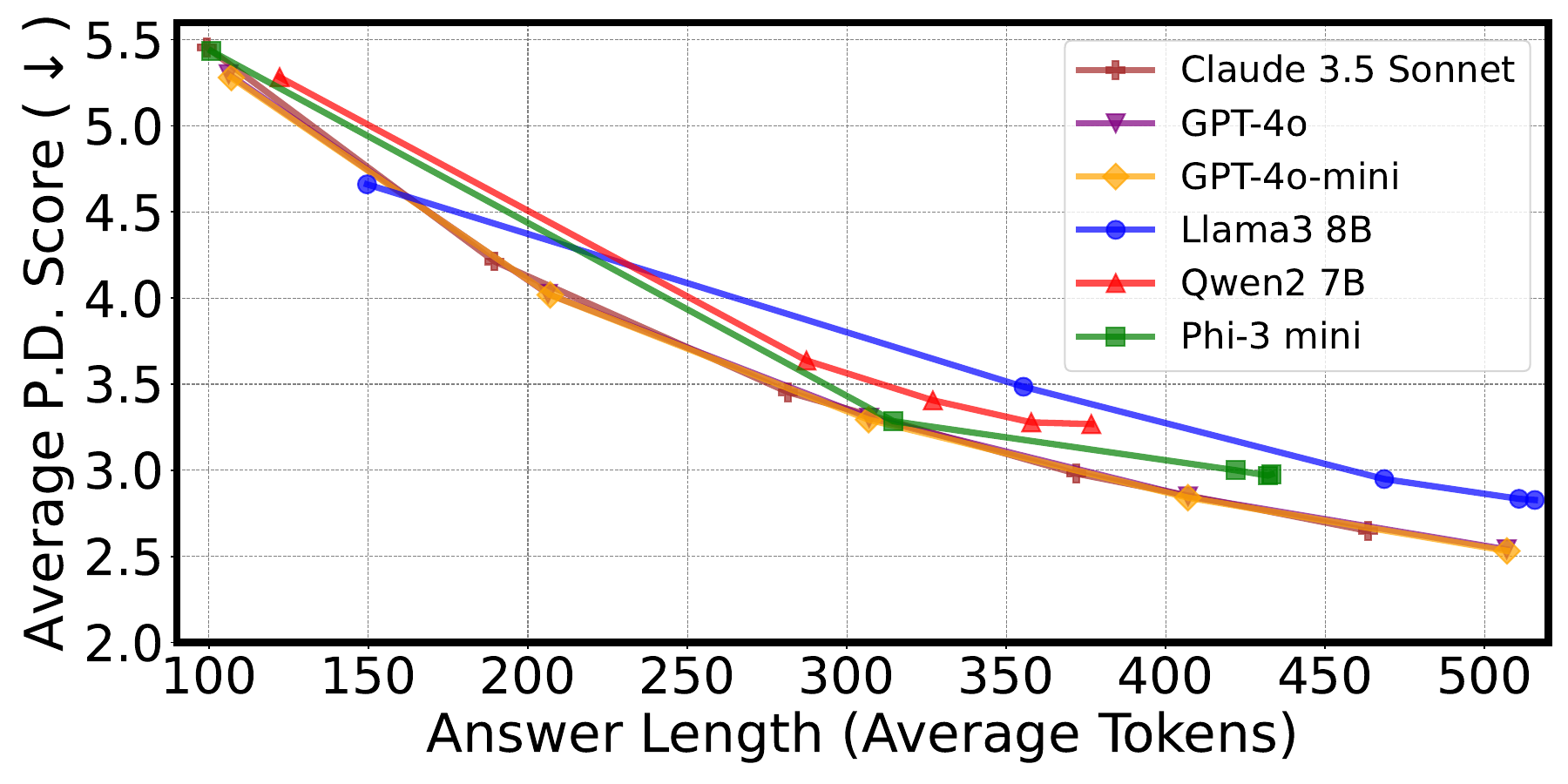}
        \label{fig:PD-length}
    }
    \\
    \vspace{-1em}
    \subfloat[D.A. Scores ($\uparrow$)]{
        \includegraphics[width=0.825\linewidth]{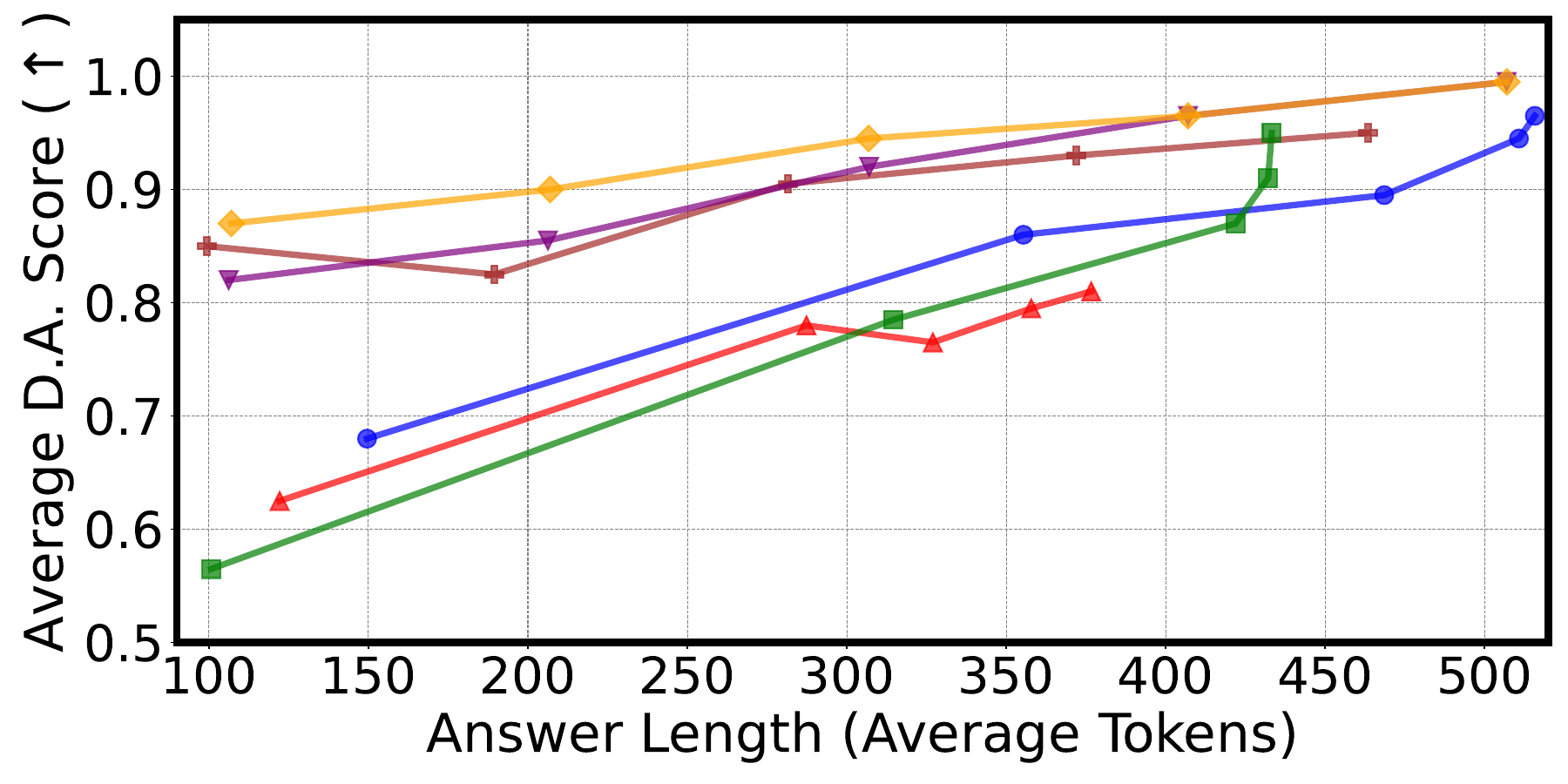}
        \label{fig:DA-length}
    }
    \caption{Average P.D. and D.A. scores for answers in various lengths across different LLMs.}
    \vspace{-1em}
    \label{fig:PD-DA-with-different-length}
\end{figure}

To delve deeper into how response length impacts the two metrics,
we use the prompt ``Your answer must be around \placeholder{num} tokens.'' to regulate LLMs to respond with a predetermined length.
However, recognizing that the open-source model's adherence to instructions might be inconsistent, we illustrate the correlation between the \emph{actual} average token count in the model's responses in~\autoref{fig:PD-DA-with-different-length}.
We find: (1) models tend to perform better with longer responses. This is likely due to longer answers providing more comprehensive information, enhancing P.D. scores. Furthermore, when tasked with longer answers, models are more prone to acknowledge the debate, which improves D.A. scores. (2) in the main experiment, GPT-4o outperforms GPT-4o-mini and Claude 3.5 Sonnet significantly, as shown in \autoref{tab:main-results}. However, the performance gap narrows when responses are constrained to equal lengths. This suggests that while the knowledge and conversational capabilities of the three models are comparable, GPT-4o's propensity for completing longer answers gives it an edge over the other two, which favor brevity.

\subsection{Qualitative Analysis}

Despite GPT-4o leading in both metrics, certain aspects of its performance remain unsatisfactory. We also examine the performance gaps among less capable models. Through case studies, our conclusions are: (1) state-of-the-art LLMs still fall short in leveraging referenced and convincing information for their arguments, and (2) weaker LLMs have difficulty providing a comprehensive elaboration of sufficient perspectives and we find the answer presentation of those LLMs are markedly poorer.
Based on these gaps, LLMs' ability to address debatable questions still has room to be further refined.
We point out the following potential solutions for improving models' ability on \dataset:
(1) \emph{Deepen domain-specific knowledge} with pretraining or high-quality RAG;
(2) \emph{Strengthen source citation} by adopting strategies such as citing memory or retrieved documents for evidence-rich responses;
(3) \emph{Improve stylistic presentation} through fine-tuning; and
(4) \emph{Ensure perspective diversity and balance} through alignment training or prompt engineering.
Kindly refer to \autoref{sec: case-studies} for a detailed analysis and case studies.

%% file: Tabs/experiment_test_models.tex
\begin{table*}[ht]
\setlength{\tabcolsep}{1pt} 
\fontsize{9}{9}\selectfont
\centering
\begin{threeparttable}
\begin{tabular}{lccccccccccc}
\toprule
\multirow{3}{*}{\textbf{Model}} &\multirow{3}{*}{\parbox{1.4cm}{\centering Avg. Len.\\(\#tokens)}} & \multicolumn{5}{c}{\criteriaone} & \multicolumn{5}{c}{\criteriatwo} \\
\cmidrule(lr){3-7} \cmidrule(lr){8-12}
&& \multicolumn{2}{c}{\llmeval{}=Qwen2 0.5B} &\multicolumn{2}{c}{\llmeval{}=GPT-2} 
&\multirow{2}{*}{\parbox{1.2cm}{\centering Norm. \\ Rank}} & \multicolumn{2}{c}{\llmeval{}=Phi-3 M.} &\multicolumn{2}{c}{\llmeval{}=Qwen2 1.5B} 
&\multirow{2}{*}{\parbox{1.2cm}{\centering Norm. \\ Rank}} \\
\cmidrule(lr){3-4} \cmidrule(lr){5-6} \cmidrule(lr){8-9} \cmidrule(lr){10-11} 
&& Score $\downarrow$ & Rank  & Score $\downarrow$  & Rank &  & Score $\uparrow$ & Rank  & Score $\uparrow$  & Rank \\
\midrule
\multicolumn{12}{c}{\emph{Closed-Source LLMs}} \\
\midrule
GPT-4o & 434& \ctext{mygreen!20}{3.07} & 1 & \ctext{mygreen!20}{4.03}& 1 & 1 &\ctext{mygreen!20}{0.952} & 1 & \ctext{mygreen!20}{0.979}& 1 & 1\\
GPT-4o mini&252 &4.09 &6 & 5.88& 6& 6 & 0.937& 4 & 0.964& 4& 4\\
GPT-3.5 Turbo &141& 5.28 & 10 & 8.25 & 10 &10& 0.904 & 6 & 0.947 & 6 &6 \\
Claude 3.5 Sonnet& 199& 4.63 & 8 & 6.96 & 8 &8  & 0.856 & 10 &0.920 & 9 &10\\
\midrule
\multicolumn{12}{c}{\emph{Open-Source LLMs (Medium to Large)}} \\
\midrule
Llama3 70B & 432 & 3.09 &2  & 4.07 & 3 &2=& 0.945 & 3  & 0.977& 2 &2=\\
Llama3 8B&381& 3.51 & 5 & 5.02 & 5 &5& 0.928 & 5 & 0.964& 4 &5\\
Qwen2 7B&255& 4.18 & 7& 6.10 & 7& 7& 0.895 & 8& 0.923& 8& 8 \\
Phi-3 small 128k& 412& 3.50 &4  & 4.31&4 & 4& 0.899 & 7  & 0.924&7 & 7\\
Gemma 2 9B &395 &3.12& 3 & 4.04 & 2 & 2=&0.947 & 2 & 0.967& 3 & 2=\\
\midrule
\multicolumn{12}{c}{\emph{Open-Source LLMs (Tiny to Small)}} \\
\midrule
Qwen2 1.5B&169& 5.60 & 11 &  8.67 & 11 & 11& 0.864 & 9 &  0.875& 10 & 9\\
Qwen2 0.5B&72& \ctext{myred!20}{6.56} & 12 & \ctext{myred!20}{10.87}& 12& 12& 0.792 & 11 & 0.836& 11& 11 \\
Phi-3 mini 128k&218 &4.82 & 9&7.33 & 9&9&\ctext{myred!20}{0.716} & 12&\ctext{myred!20}{0.794} &12 &12 \\
\bottomrule
\end{tabular}
\end{threeparttable}
\caption{\label{tab:main-results} 
Main results of P.D. and D.A. for LLMs on \dataset{}-\texttt{test}. Avg. Len.: average length of the answers, GPT-2: GPT-2 (117M), Phi-3 M.: Phi-3 medium 128k, Norm. Rank: normalized average rank of different \llmeval{}. The \ctext{mygreen!20}{best} and \ctext{myred!20}{worst} results of each metric (\wrt{} a specific \llmeval{}) are highlighted.
}
\end{table*}

%% file: Secs/conclusion.tex
\section{Conclusion}
\label{sec:conclusion}

We develop \dataset{}, a novel QA dataset with 2,941 debatable questions paired with multiple human-annotated partial answers to assess language models' ability to answer debatable questions. We introduce two novel metrics, Perspective Diversity and Dispute Awareness, to evaluate how well language models reflect multiple human viewpoints and recognize debate. Extensive experiments demonstrate these two metrics align with human preferences and are stable. Our evaluations with 12 LLMs show that current models are relatively strong in recognizing debate but varying in the ability to comprehensively address them.

%% file: Appens/dataset.tex
\section{Further Details on Dataset Curation}
\label{sec: dataset-appen}

\subsection{Details on Debatable Question}
\label{subsec: sources_and_deduplication}

The questions in \dataset{} are collected from three sources, as detailed in~\autoref{tab:dataset-distribution-sources}.

\input{Tabs/dataset_sources}

\begin{table}[ht]
\fontsize{8.5}{8.5}\selectfont
\centering
\begin{threeparttable}
\begin{tabular}{lcc}
\toprule
\textbf{Sources} &Initial &After Dedupe.   \\
\midrule
DELPHI&2,281 &1,597 \\
\textsc{ConflictingQA}&434 &400 \\
Web Sourced&1,758 &1,219 \\
\midrule
\textbf{Total} &4,473 &3,216 \\
\bottomrule
\end{tabular}
\end{threeparttable}
\caption{\label{tab:dataset-distribution-sources} Sources distribution of \dataset{}.}
\end{table}

\noindent \textbf{Manually sourcing debatable questions.}
We search for debatable questions from the Web and adhere to the following criteria:
\begin{packeditemize}
\item A debatable question should have clear semantics without any ambiguity.
\item A debatable question should have potentially different angles of answer, rather than having one definitive answer.
\item A debatable question can be subjective, but its answer should not be determined \emph{solely} by individual subjective preferences.
\end{packeditemize}
In the end, we select 1,758 debatable questions from 9 websites. Details on the number of questions from each website can be found in~\autoref{tab:dataset-sources}.

\noindent \textbf{Question deduplication.}
After merging questions from the three sources, we employ a simple deduplication algorithm to remove duplicates from the original set of 4,473 questions:
First, we calculate cosine similarities between pairwise questions based on the embeddings computed by gte-large-en-v1.5~\cite{DBLP:journals/corr/abs-2308-03281}. Subsequently, we sample 500 pairs and find pairs with a similarity score below 0.78 contained no true duplicates.
Following this, we leverage a greedy algorithm for minimum vertex cover to remove duplicate entries, aiming to remove as few questions as possible while ensuring no duplicates remain\footnote{The problem of vertex cover is NP-hard. We leverage a greedy algorithm, similar to the one described in \href{https://courses.cs.washington.edu/courses/cse421/21sp/vertex-cover.pdf}{this page}.}. This process results in a final dataset consisting of 3,216 unique entries.

\subsection{Details on Retrieving Trustworthy Documents}
\label{subsec: retrieve-documents-appen}

\noindent \textbf{Retrieving on trustworthy websites.} We only do retrieval on authoritative domains in~\autoref{tab:TLDs} to assure the trustworthiness of the documents.
Among the selected TLDs, \texttt{.gov} and \texttt{.edu} domains are not open for personal registration and can only be registered by government or educational institutions. Although \texttt{.org}, \texttt{.pro}, and .\texttt{info} domains can now be registered by individuals, their content generally remains professional and informative, with fewer advertisements or potentially misleading information. 

\begin{figure}[ht]
    \centering
    \includegraphics[width=0.8\linewidth]{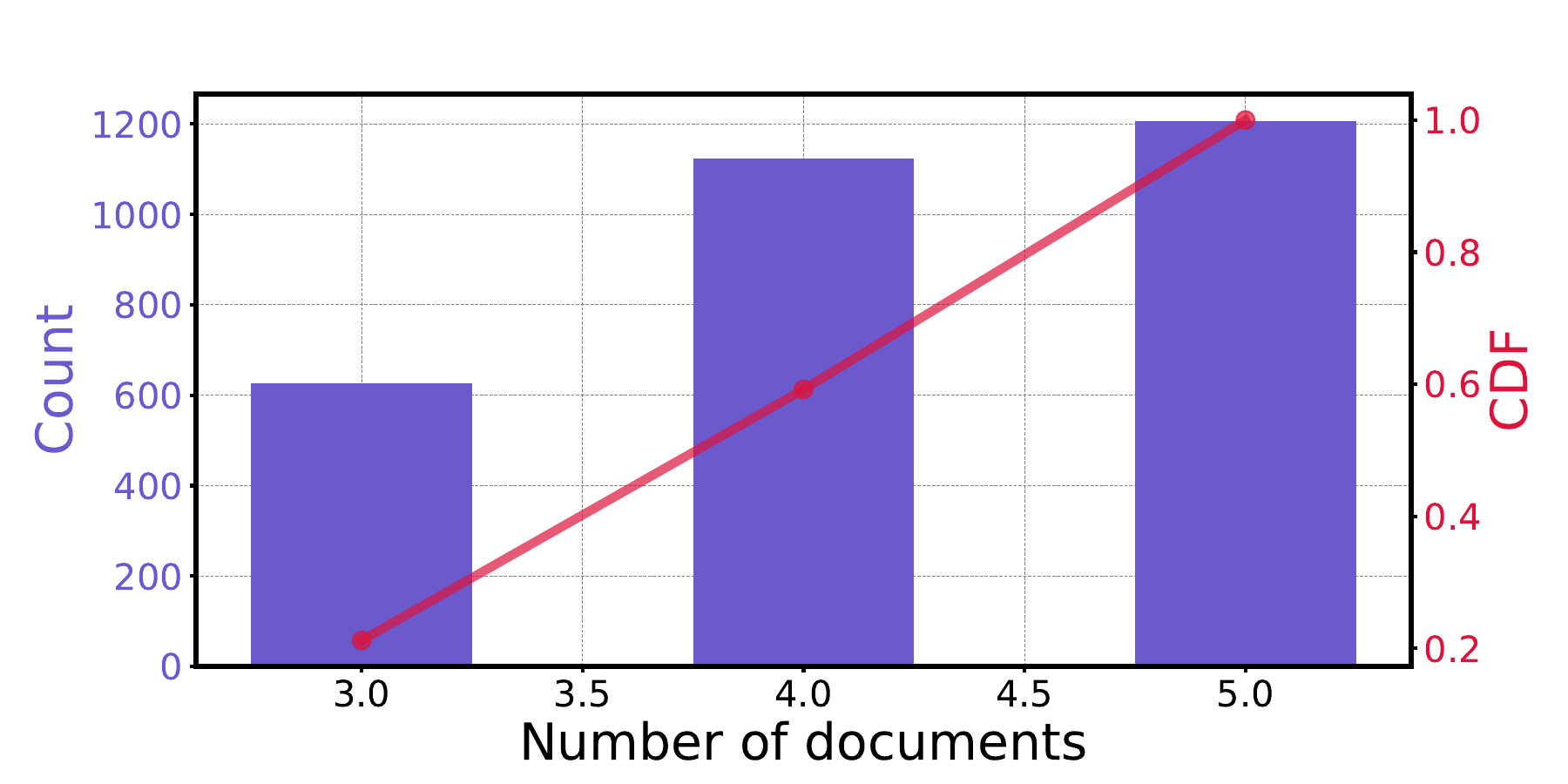}
    \vspace{-1em}
    \caption{Distribution of the number of trustworthy evidence documents per question.}
    \vspace{-1em}
    \label{fig:distribution-num-documents}
\end{figure}

\noindent \textbf{Implementation of the retrieving process.}
To enable finer-grained search results, we apply the GPT-4 model to first transform the original question into several search queries.
We use the \href{https://developers.google.com/custom-search/v1/introduction}{Google search engine} for Web searches and retain only the documents from authoritative TLDs. These documents are then ranked using Bge-Reranker-v2-Gemma~\cite{chen2024bge} and we keep the top-5 documents.
We filter questions with fewer than three documents, as we consider these lack sufficient trustworthy evidence, leaving us with 2,982 questions. The distribution of the number of documents per question is in~\autoref{fig:distribution-num-documents}.

\begin{figure}[ht]
    \centering
    \includegraphics[width=0.9\linewidth]{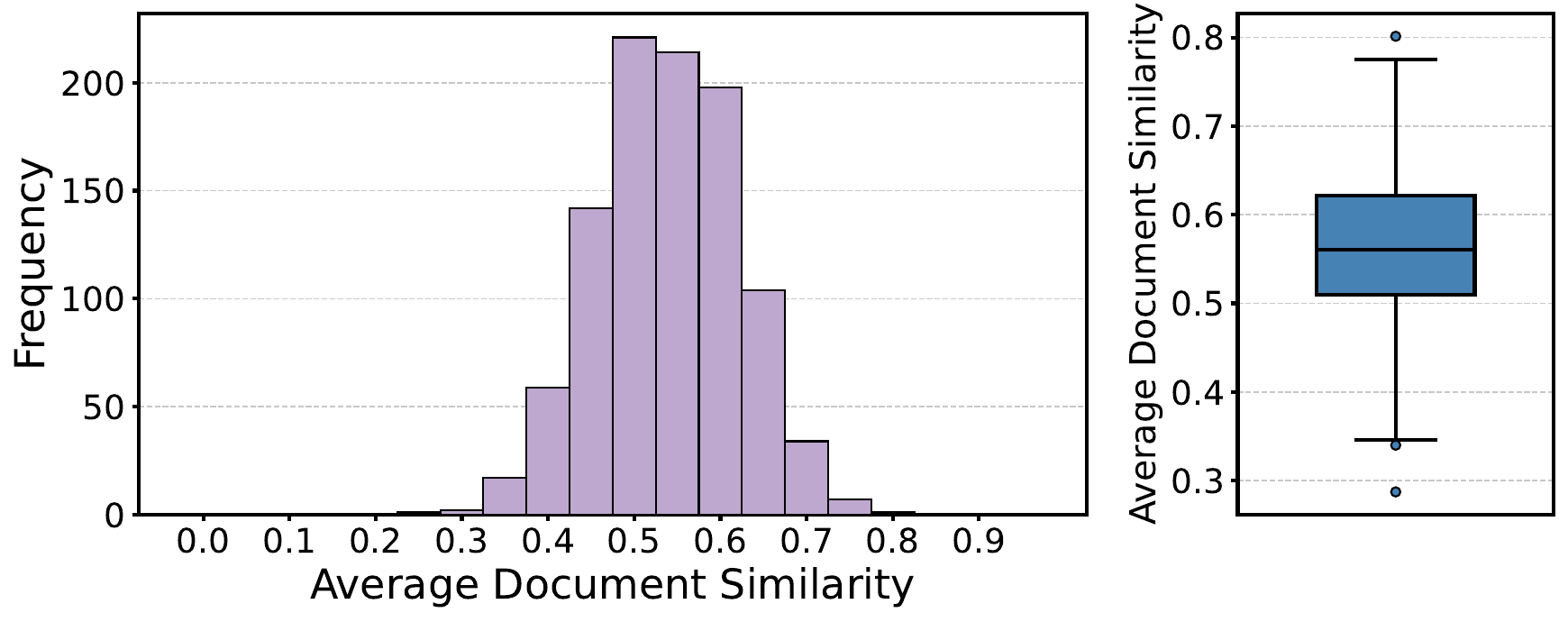}
    \vspace{-1em}
    \caption{Quality of retrieved evidence documents. Document quality is assessed by cosine similarity.}
    \vspace{-2em}
    \label{fig:3a-document-similarity}
\end{figure}

\noindent \textbf{Quality examination.}
We segment each document into 1000-token chunks and average the cosine similarities for each question and corresponding trunks, computed by gte-Qwen2-1.5B-instruct. 
The quality of the retrieved documents is illustrated in \autoref{fig:3a-document-similarity}.

\begin{table}[ht]
\fontsize{8.5}{8.5}\selectfont
\centering
\begin{threeparttable}
\begin{tabular}{cc}
\toprule
\textbf{Selected TLDs}  & Open for Registration? \\
\midrule
\texttt{.edu} & \crossmark\\
\texttt{.org} &  \checkmark\\
\texttt{.gov} & \crossmark\\
\texttt{.info}&  \checkmark\\
\texttt{.pro} & \checkmark\\
\bottomrule
\end{tabular}
\end{threeparttable}
\caption{\label{tab:TLDs} List of selected top-level domains (TLDs) we considered trustworthy. Closed for registration indicates that only authoritative entities can register a domain under these TLDs, ensuring high credibility.}
\vspace{-1em}
\end{table}

\subsection{Details on Generating Partial Answers}
\label{subsec: generating-partial-answers-appen}

\autoref{alg:partial-answers} formalize the pipeline of collecting partial answers, where $\mathcal{M}$ is the LLM we use. $\mathcal{M}(p(x,y))$ indicates the LLM processing a prompt template $p()$ populated with inputs $x,y$.

\begin{algorithm}[ht]
\caption{Collecting partial answers for an individual debatable question}
\fontsize{9}{9}\selectfont
\label{alg:partial-answers}
\DontPrintSemicolon
\SetNoFillComment
\KwData{question $q$, a list of $m$ evidence documents $\mathcal{D} = \{D\}_{i=1}^m$}
\KwResult{a list of $n$ partial answers $\mathcal{P} = \{P_i\}_{i=1}^n = \{(\text{POV}_i, \text{Explan}_i)\}_{i=1}^n$}
\tcc{Extract list of POVs\\ $\mathcal{D}_{\text{rel.}}i$ are relevant docs \wrt{} $\text{POV}_i$}
$\{\text{POV}\}_{i=1}^n, \{\mathcal{D}_{\text{rel.}}\}_{i=1}^n \gets \mathcal{M}(p_{\text{POV}}(q, \mathcal{D}))$\\
\For{\em $\text{POV}_i$ in $\{\text{POV}\}_{i=1}^n$}{
\tcc{Expand POV to Explanations}
$\text{Explan}_i \gets \mathcal{M}(p_{\text{Explan}}(q, \text{POV}_i, \{\mathcal{D}_{\text{rel.}}\}_i))$
}
\Return $\mathcal{P} \gets \{(\text{POV}_i, \text{Explan}_i)\}_{i=1}^n$
\end{algorithm}

\begin{figure}[ht]
    \centering
    \includegraphics[width=0.8\linewidth]{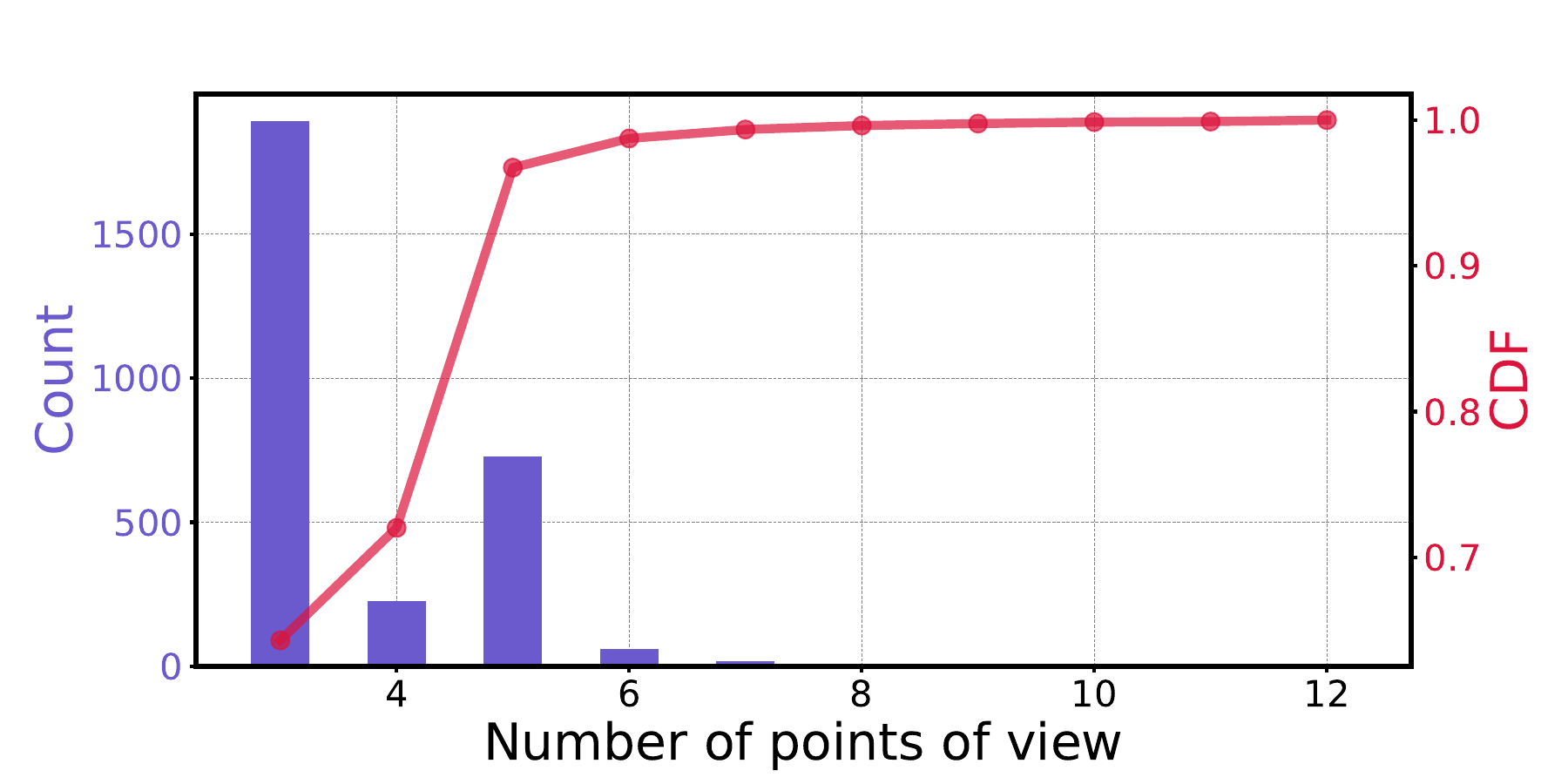}
    \vspace{-1em}
    \caption{Distribution of the number of extracted POVs per question.}
    \label{fig:distribution-num-POVs}
    \vspace{-1em}
\end{figure}

\noindent \textbf{Prompts.}
The prompts we used to generate the POVs and explanations can be found in \autoref{tab:core-prompts}.
These prompts are carefully crafted to ensure that the generated POVs cover a range of non-overlapping perspectives and provide well-rounded explanations that are grounded in the evidence documents.
After extracting the POVs, we filter out questions with fewer than three perspectives, ensuring that the remaining questions are sufficiently debatable, resulting in 2,941 questions.
The distribution of the number of extracted POVs per question can be found in~\autoref{fig:distribution-num-POVs}.

\noindent \textbf{Configuration of the LLM.}
At the time of dataset curation (from April to May 2024), the strongest model available was the \texttt{gpt-4-turbo} variant of GPT-4. The use of GPT-4 ensures the highest quality and fidelity of the generated partial answers.
We set top-$p=0.7$ to enable nucleus sampling~\cite{DBLP:conf/iclr/HoltzmanBDFC20} and $\text{temperature}=0.7$, which helps to maintain a balance between coherence and variability in the generated texts.

\noindent \textbf{Preprocessing the documents.}
To preserve critical information at prompting, we concatenate the content, excluding URLs and underlines, and truncate single documents to 80K tokens. If the concatenated input exceeds 120K tokens, we trim to this limit, ensuring essential content fits within GPT-4's 128K context window.

\noindent \textbf{Quality examination.}
We examine the quality of POVs based on two criteria: 
their comprehensiveness \wrt{} the retrieved documents and their distinctiveness from one another.

\noindent \emph{Comprehensiveness}: 
To ensure the comprehensiveness of extracted POVs, we examine whether all valid perspectives from the questions' corresponding documents are included in the POVs. 
To this end, we truncate the documents into 1000-token chunks and use gte-Qwen2-1.5B-instruct to identify the top 5 most relevant chunks per question. 
A manual examination reveals that in 90.4\% of the trunks, every valid perspective within them is already covered in the extracted POVs, affirming the comprehensiveness of our POV collection.

\noindent \emph{Distinctiveness}: 
To ensure the distinctiveness of extracted POVs, \ie, they are non-overlapping, we calculate the pairwise cosine similarity for each question's POVs using gte-Qwen2-1.5B-instruct. 
We set a 0.75 cosine similarity threshold to discern unique POVs. Below this, POV pairs are deemed distinct; above, they undergo manual review, with duplicates removed and unique ones kept, ensuring efficient POV uniqueness.

\input{Prompts/core_prompts}

\subsection{Details on Human Annotation}
\label{subsec: human-annotation-appen}

\begin{figure}[ht]
    \centering
    \includegraphics[width=\linewidth]{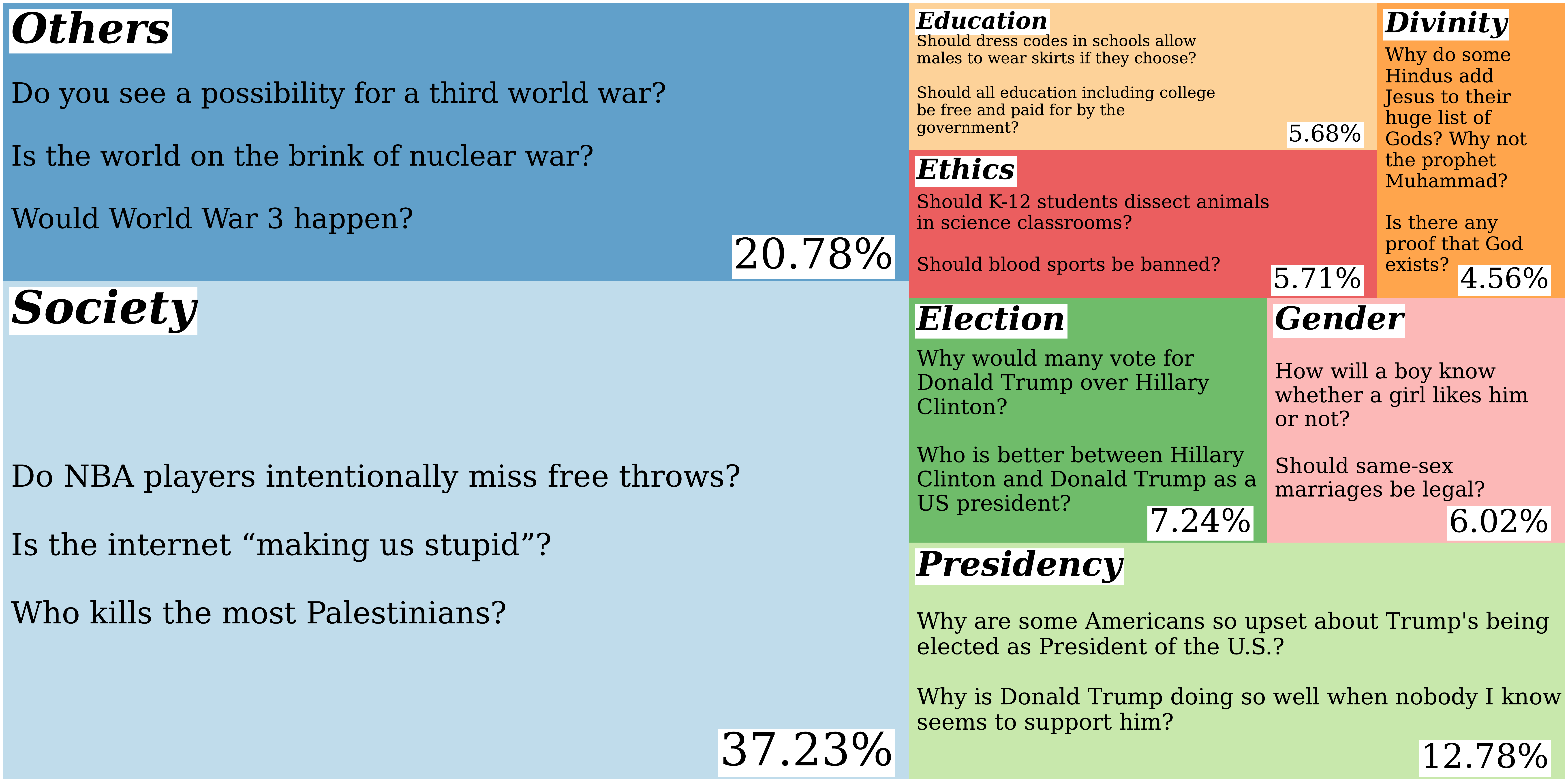}
    \vspace{-2em}
    \caption{Question distribution for top-8 domains, rendered by BERTopic~\cite{DBLP:journals/corr/abs-2203-05794}.}
    \label{fig:dataset-distribution-domain}
    \vspace{-1em}
\end{figure}

We recruit three professional annotators from a local data annotation company to verify the partial answers. The payment for this job is above the local minimum wage. 
Annotators are given two distinct tasks as outlined in \autoref{subsec: annotation-dataset}. These tasks involve making binary decisions, where annotators must assess if the partial answer satisfies the specified criteria in \autoref{subsec: annotation-dataset}.
After the annotation, we removed 767 partial answers deemed substandard by two or more annotators, resulting in a final dataset of 10,873 partial answers. 
We do not remove the original questions corresponding to these partial answers, as those questions still have multiple partial answers.
A domain distribution of the final dataset is shown in~\autoref{fig:dataset-distribution-domain}.

%% file: Tabs/dataset_sources.tex
\begin{table*}[ht]
\fontsize{8}{8}\selectfont
\setlength{\tabcolsep}{1pt} 
\centering
\begin{threeparttable}
\begin{tabularx}{\linewidth}{Xc}
\toprule
\textbf{Source URL} &Count\\
\midrule
\url{https://owlcation.com/academia/debate-topics} & 350 \\
\url{https://paperperk.com/blog/debate-topics} & 174 \\
\url{https://studycorgi.com/blog/debatable-questions-topics-for-research-essays/} &  150 \\
\url{https://www.myspeechclass.com/funny-debatable-topics.html} & 126 \\
\href{https://www.nytimes.com/2020/07/28/learning/177-questions-to-inspire-writing-discussion-debate-and-reflection.html}{\texttt{https://www.nytimes.com/2020/07/28/learning/177-questions...refection.html}} & 177 \\ 
\url{https://owlcation.com/academia/100-Debate-Topics} & 100 \\
\url{https://parade.com/living/debatable-questions} & 257 \\
\url{https://www.procon.org/debate-topics} & 107 \\
\url{https://randomquestionmaker.com/blog/debatable-questions} & 317 \\
\bottomrule
\end{tabularx}
\end{threeparttable}
\caption{\label{tab:dataset-sources}
Detailed sources in URL for debatable questions sourced from the web. During our experiment (April 2024), these resources are accessible. We will make our dataset publicly available for future research.
}
\vspace{-1em}
\end{table*}

%% file: Prompts/core_prompts.tex
\begin{table*}[ht]
\fontsize{7.5}{7.5}\selectfont
\setlength{\tabcolsep}{2pt} 
\centering
\begin{threeparttable}
\begin{tabularx}{\linewidth}{lX}
\toprule
\textbf{Prompt} & Content \\
\midrule
$p_{\text{POV}}$ &    
   Task: Generate Points-of-View that Address the Given Question from Different Perspectives\newline
Guidelines:\newline
1. Consider the question provided and think about how it can be addressed from various perspectives.\newline
2. Use the information from the provided documents. Do not rely on your internal knowledge.\newline
3. Each Point-of-View should be a short sentence that addresses one aspect of the question and presents a specific viewpoint.\newline
4. Ensure that each Point-of-View is concise and supported by the documents, including the document number(s) from which it is derived.\newline
5. Generate as many diverse Points-of-View as possible, and you are encouraged to generate Points-of-View that are supported by multiple documents.\newline
Question: \placeholder{question} \newline 
Documents: \placeholder{concatenated documents}\newline
Develop Points-of-View by drawing insights from the text, allowing each one to present a specific perspective. \newline
Format your response as follows: start each Point-of-View with ``Point-of-View [number]:'', followed by its content, and include a list of document number(s) related to that Point-of-View.\newline
Expected Output Format:\newline
Point-of-View 1: (content of the Point-of-View) [Document [number]]\\
\midrule
$p_{\text{Explan}}$ &   
Task: Provide an Answer to a Question that Reflects a Given Perspective\newline
Guidelines: \newline 1. Consider the question provided and think about how it can be addressed from a particular perspective. \newline 2. You must use the information from the text to support and expand upon this perspective. You must not rely on your internal knowledge. \newline 3. Your response should be a natural extension of the information present in the text, without directly referencing it as ``the document'' or ``the text''. \newline 4. Your response should include at most 300 words.\newline 
Question: \placeholder{question} \newline 
Perspective: \placeholder{POV} 
\newline Text: \placeholder{(concatenated) document(s) \wrt{} a specific POV} \newline Develop your answer by drawing insights from the text, allowing the answer to unfold as a natural expansion of the provided perspective.\\
\bottomrule
\end{tabularx}
\end{threeparttable}
\caption{\label{tab:core-prompts}
Core prompts for generating partial answers given the question and retrieved documents.
Prompt $p_{\text{POV}}$ is used to extract points-of-view (POVs) from the corresponding evidence documents \wrt{} to the question. 
Prompt $p_{\text{Explan}}$ is used to expand POVs into long-form explanations based on the relevant documents.
}
\vspace{-1em}
\end{table*}

%% file: Appens/metric_PD.tex
\section{Further Details on P.D.}
\label{sec:PD-appen}

\subsection{P.D.'s Alignment with Human Preferences}
\label{subsubsec:PD-llms}

\subsubsection{Collecting Human Preferences}
\label{subsubsec:annotation-preferences}

Initially, we gather responses from a diverse selection of five LLMs to 100 randomly chosen test questions from \dataset{}-\texttt{test}. The list of LLMs is as follows: 
GPT-4o~\cite{openai2024gpt4o}, Llama 3 70B~\cite{meta2024llama3}, Phi-3 Small 8k~\cite{DBLP:journals/corr/abs-2404-14219}, Zephyr 7B beta~\cite{DBLP:journals/corr/abs-2310-16944}, and Qwen1.5 4B~\cite{qwenteam2024qwen15}, representing a range of manufacturers and capabilities, anticipated to produce varying response qualities. 
We configure the LLMs as described in~\autoref{subsec: experimental-setup} to solicit answers, resulting in 500 answers.
Subsequently, we engage three annotators to record their preferences among the model answers.
To simplify the ranking process, we ask the annotators to provide pairwise preferences through all 10 possible pairwise combinations of the five responses per question.
The annotators need to provide a preference based on the following criteria:

\begin{packeditemize}
    \item Clearly indicate that the question being answered is controversial (possibly without a universally accepted answer).
    \item Try to comprehensively cover various angles of the controversial issue.
    \item For each viewpoint in the answer, use specific, sufficient, credible, and supportive evidence to elaborate.
    \item Try to avoid letting your subjective understanding of the issue affect your choice of answer.
\end{packeditemize}

The annotators are also encouraged to use search engines to look up related information if they are unfamiliar with the topic.

Post-annotation, we determine inter-annotator consistency using Spearman's $\rho$ correlation. The outcomes, depicted in \autoref{fig:correlation-preference}, reveal strong agreement ($\rho>0.8$) among annotators, suggesting a shared understanding of a \textbf{good} answer.

\subsubsection{Baseline Text Evaluation Metrics}
\label{subsubsec:baseline-evaluation}

The prompts for P.D.'s baseline metrics can be found \autoref{tab:baseline-prompts}.

\noindent \textbf{Direct-Score}. Direct-Score is \textbf{basic} prompt-based evaluation metric. We employ a straightforward prompt that requires the model to assign a 1-5 Likert scale score to the model response using the \textbf{same} instruction we present to human annotators. The prompt $p_{\text{DS}}$ is depicted in~\autoref{tab:baseline-prompts}. 

\noindent \textbf{G-Eval}~\cite{DBLP:conf/emnlp/LiuIXWXZ23}. G-Eval is a \textbf{strong} prompt-based evaluation framework that assesses the quality of generated texts by incorporating chain-of-thoughts (CoT)~\cite{DBLP:conf/nips/Wei0SBIXCLZ22} and a form-filling paradigm. By providing a prompt with a task introduction and evaluation criteria, G-Eval generates detailed evaluation steps and utilizes these steps along with the generated CoT to score the texts. We apply G-Eval using the \textbf{same} scoring criteria provided to humans. 
The prompt $p_{\text{G-Eval}}$ behind G-Eval can be found in~\autoref{tab:baseline-prompts}.

\noindent \textbf{Num-of-POVs}. 
We design another prompt-based evaluation metric that takes a \textbf{shortcut} approach by simply determining the number of different perspectives in an answer. This metric can be considered an improved metric over the ``Comprehensiveness Answer Rate'' metric introduced in the DELPHI paper, as it transcends the binary assessment of the original, which solely determines if an answer includes diverse and opposing viewpoints. The prompt $p_{\text{NoP}}$ is shown in~\autoref{tab:baseline-prompts}. 

For the \textbf{similarity-based metrics}, the score for an answer is calculated by averaging the individual scores computed with the partial answers.

\noindent \textbf{BLEU}~\cite{DBLP:conf/acl/PapineniRWZ02} and \textbf{ROUGE-L}~\cite{lin2004rouge}. These are statistical metrics. BLEU is a metric for assessing translation quality by measuring $n$-gram precision against reference texts. 
ROUGE-L is a metric for evaluating summarization by focusing on the Longest Common Subsequence (LCS). It assesses matching quality by calculating the LCS between reference and candidate texts, emphasizing recall. 

\noindent \textbf{BARTScore}~\cite{NEURIPS2021_e4d2b6e6} and \textbf{SemScore}~\cite{DBLP:journals/corr/abs-2401-17072}. These are neural metrics. BARTScore evaluates the quality of text by leveraging the generative capabilities of the BART~\cite{DBLP:conf/acl/LewisLGGMLSZ20} model. 
We employ the \emph{Precision} score, which measures the likelihood of a generated answer (candidate text) matching a given reference answer (reference text). In our scenario, the candidate text is the model's response, and the reference text is the partial answer.
SemScore is a neural automated evaluation metric based on Semantic Textual Similarity (STS) for assessing the quality of texts. 

\input{Prompts/P.D._baseline_prompts}

\subsubsection{Case Study}
\label{subsubsec:PD-casestudy}

Despite P.D.'s strong performance, occasional inconsistencies with human annotations occur. We delve into these by analyzing one set of answers with perfect P.D.-human alignment and three divergent sets. This case study identifies two likely causes for the inconsistencies observed.

\noindent\textbf{Non-perspective content in model's response.} The first reason is that the answers provided by LLMs may contain segments that cannot be directly considered as a POV or arguments that support a POV \wrt{} the question. For instance, when answering the question ``Should students have to wear school uniforms?'', Llama 3 70B's answer includes a sentence ``What do you think? Should students have to wear school uniforms, and why or why not?'', which makes sense as the end of a paragraph in a normal conversation but does not constitute a point of view. Consequently, it does not appear in the partial answers in our dataset. Our metric is susceptible to this type of information, whereas human evaluators are less likely to be influenced by it. This results in humans rating this answer higher than our metric does.

\noindent\textbf{Omission of legitimate perspectives.} The second reason is that although our partial answer has covered many credible perspectives on our debatable question, some perspectives may still have been missed given that we only extract perspectives from the top documents of a Google search. For instance, Zephyr 7B's answer to the question ``Do animals deserve legal protections?'' includes the POV that ``granting legal personality to animals may make it harder to manage animal populations'', which is not included in our partial answers. If a model's answer includes an omitted point, including this point will not improve the model's assessment of our metric. However, human annotators do not have a \emph{predetermined pool of perspectives}, so including this point will likely make the annotator think the answer is more comprehensive, leading to discrepancies in judgments.

\subsection{P.D. is Stable \wrt{} Backbone Models}

We choose the following language models: Qwen2 0.5B/7B~\cite{qwen2023qwen2}, Phi-3 mini 128k~\cite{DBLP:journals/corr/abs-2404-14219}, GPT-2 (117M)~\cite{radford2019language}, MiniCPM 2B~\cite{DBLP:journals/corr/abs-2404-06395}.

\subsection{P.D. is Stable \wrt{} Prompts}

As in \autoref{tab:PD-stability-prompts}, we use five different prompts to show that P.D. is stable \wrt{} prompts.

\input{Prompts/P.D._stability_prompts}

%% file: Prompts/P.D._baseline_prompts.tex
\begin{table*}[ht]
\fontsize{7.5}{7.5}\selectfont
\setlength{\tabcolsep}{2pt} 
\centering
\begin{threeparttable}
\begin{tabularx}{\linewidth}{lX}
\toprule
\textbf{Prompt} & Content \\
\midrule
$p_{\text{DS}}$ & You will be given a question and its answer. You are supposed to give a score to represent the quality of the answer. \newline
    A high-quality answer should meet the following criteria: \newline
    1. Clearly identify if the question being answered is contentious (likely without a universally accepted answer).\newline
    2. Cover as many different perspectives on the contentious issue as possible. The more perspectives, the better.\newline
    3. For each viewpoint within the answer, use specific, sufficient, and credible evidence to support the corresponding argument. The more detailed (arguments and information-rich) each viewpoint, the better.\newline
    4. When making your judgment, prioritize the quality of the answer. Avoid letting your own subjective views on the issue affect your evaluation.\newline
    Question: \placeholder{question} \newline
    Answer: \placeholder{answer} \newline
    Your response should only contain one integer number between 1-5 to represent the quality of the answer without any other symbols or words. \\
\midrule
$p_{\text{G-Eval}}$ &     You will be given a question and its answer. Your task is to rate the answer based on its quality. \newline
    Evaluation Criteria:  \newline
    Quality (1-5) - The answer should meet the following criteria:  \newline
    1. Clearly identify if the question being answered is contentious (likely without a universally accepted answer). \newline
    2. Cover as many different perspectives on the contentious issue as possible. The more perspectives, the better. \newline
    3. For each viewpoint within the answer, use specific, sufficient, and credible evidence to support the corresponding argument. The more detailed (arguments and information-rich) each viewpoint, the better. \newline
    4. When making your judgment, prioritize the quality of the answer. Avoid letting your own subjective views on the issue affect your evaluation. \newline
    Evaluation Steps:  \newline
    1. Read the question carefully.  \newline
    2. Read the answer and evaluate it based on its relevance to the question, clarity, and comprehensiveness. \newline
    3. Assign a score for quality on a scale of 1 to 5, where 1 is the lowest and 5 is the highest based on the Evaluation Criteria. \newline
    Question: \placeholder{question}  \newline
    Answer: \placeholder{answer}  \newline
    Evaluation Form (output one number to represent scores ONLY):  \newline
    - Quality:
    \\
\midrule
$p_{\text{NoP}}$ &     You will be given a question and its answer. 
    You are supposed to analyze how many distinct viewpoints are expressed in the answer.\newline
    Your response must be a single number such as 1, 2, etc. \newline
    Your answer should only contain a single Arabic numeral between 1 and 10 without any other symbols or words. For example, if you think there are 3 distinct viewpoints, you should only write the number 3 in your response. \newline
    Question: \placeholder{question} \newline
    Answer: \placeholder{answer} 
    \\
\bottomrule
\end{tabularx}
\end{threeparttable}
\caption{\label{tab:baseline-prompts}
Prompts for assessing answers to debatable questions in baseline methods of the P.D. metric.
}
\vspace{-1em}
\end{table*}

%% file: Prompts/P.D._stability_prompts.tex
\begin{table}[ht]
\fontsize{8}{8}\selectfont
\setlength{\tabcolsep}{2pt} 
\centering
\begin{threeparttable}
\begin{tabularx}{\linewidth}{lX}
\toprule
\textbf{Prompt} & Content \\
\midrule
$p_{\text{P.D. 1}}$ &  \placeholder{response}   Please restate.  
    \\
\midrule
$p_{\text{P.D. 2}}$ & Here is the text:  \placeholder{response}   What is described in the text?    
    \\
\midrule
$p_{\text{P.D. 3}}$ &  Given the following text:  \placeholder{response}  Please rephrase the text.    
    \\
\midrule
$p_{\text{P.D. 4}}$ &  From the following text:  \placeholder{response}   What we can infer?   
    \\
\midrule
$p_{\text{P.D. 5}}$ &     \placeholder{response}
    \\
\bottomrule
\end{tabularx}
\end{threeparttable}
\caption{\label{tab:PD-stability-prompts}
Prompts for assessing answers to debatable questions in baseline methods of the P.D. metric.
}
\vspace{-1em}
\end{table}

%% file: Appens/metric_DA.tex
\section{Further Details on D.A.}
\label{sec:DA-appen}

The exact prompt for D.A. is shown in \autoref{tab:DA-prompts}.

\input{Prompts/D.A._prompts}

\subsection{D.A. is Accurate}
\label{subsec:DA-accuracy-appen}

We compare the D.A. metric with the one from DELPHI~\cite{DBLP:conf/emnlp/SunAKKXW23}, which is our sole baseline due to this is the only comparable study to our knowledge. 
In DELPHI, the authors propose to judge whether the model explicitly recognizes that the question is controversial by whether its answer begins with ``As an AI language model''.

\subsection{D.A. is Stable \wrt{} Backbone Models}

We choose the following instruction-following LLMs: Gemma 2 9B~\cite{google2024gemma}, Llama3 8B~\cite{meta2024llama3}, MiniCPM 2B~\cite{DBLP:journals/corr/abs-2404-06395}, Phi-3 medium 128k~\cite{DBLP:journals/corr/abs-2404-14219}, Qwen2 1.5B~\cite{qwen2023qwen2}.

%% file: Prompts/D.A._prompts.tex
\begin{table*}[ht]
\fontsize{7.5}{7.5}\selectfont
\setlength{\tabcolsep}{2pt} 
\centering
\begin{threeparttable}
\begin{tabularx}{\linewidth}{lX}
\toprule
\textbf{Prompt} & Content \\
\midrule
$p_{\text{D.A.}}$ &    
    Here is a question paired with an answer. Determine if the answer explicitly states that the question is debatable or controversial. \newline
    \textcolor{mygreen}{\textbf{5-Shot Demonstrations:}  \newline
    Question: Is it ethical to use animals for scientific research? \newline
    Answer: The use of animals in scientific research is a highly debated topic. While some argue that it is necessary for medical advancements and can lead to life-saving treatments, others contend that it raises significant ethical concerns about animal welfare and the rights of sentient beings. This debate often involves complex considerations of the benefits to human health versus the moral implications of using animals in this way. \newline
    Response: 1 \newline
    \ldots \newline
    Question: Should schools replace physical education with computer coding classes? \newline
    Answer: Schools are increasingly incorporating coding into their curriculums to prepare students for the digital age, but this does not necessarily mean that physical education should be replaced. \newline
    Response: 0} \newline
    Instruction: \newline
    - Respond with "1" if the answer explicitly states that the question is controversial or open to debate. \newline
    - Respond with "0" if the answer does not acknowledge any debate or controversy regarding the question. \newline
    - Please only consider whether there is an explicit statement regarding the controversy, do not judge on other aspects of the answer, e.g., quality, truthfulness. \newline
    Question: \placeholder{question}  \newline
    Answer: \placeholder{answer}
     
    Important: Your answer should only contain one digit 0 or 1.
    \\
     
\bottomrule
\end{tabularx}
\end{threeparttable}
\caption{\label{tab:DA-prompts}
Prompt for the D.A. metric. $p_{\text{D.A.}}$ determine if an answer explicitly recognizes the debatable nature of the question. $p_{\text{D.A.-ZS}}$ is the zero-shot version of $p_{\text{D.A.}}$ where the \textcolor{mygreen}{in-context demonstrations} are completely omitted.
}
\vspace{-1em}
\end{table*}

%% file: Appens/experiments.tex
\section{Further Details on Experiments}
\label{sec: experiment-appen}

\begin{table}[ht]
\setlength{\tabcolsep}{2pt}
\fontsize{7.5}{7.5}\selectfont
\centering
\begin{threeparttable}
\begin{tabularx}{\linewidth}{lX}
\toprule
\textbf{Role} & Prompt    \\
\midrule
system & Your task is to provide an answer to the question. \\
\midrule
user & \placeholder{question} \\
\bottomrule
\end{tabularx}
\end{threeparttable}
\caption{\label{tab:prompt-no-rag} Prompt for test vanilla LLMs.}
\vspace{-1em}
\end{table}

\begin{table}[ht]
\setlength{\tabcolsep}{2pt}
\fontsize{7.5}{7.5}\selectfont
\centering
\begin{threeparttable}
\begin{tabularx}{\linewidth}{lX}
\toprule
\textbf{Role} & Prompt   \\
\midrule
system & Your task is to provide an answer to the question. You can refer to the information provided in the \#\# Document to help you answer better, or you can leave it out. \\
\midrule
user & \#\# Document: \newline
    \placeholder{content of chunk 1} \newline
    \placeholder{content of chunk 2} \newline
    \ldots \newline
    \placeholder{content of chunk $n$} \newline
    \#\# Question: \placeholder{question} \\
\bottomrule
\end{tabularx}
\end{threeparttable}
\caption{\label{tab:RAG-prompts} Prompt for test LLMs w. RAG.}
\vspace{-2em}
\end{table}

\subsection{RAG Configuration Details}
\label{subsec: rag-config-appen}

\noindent\textbf{Vanilla RAG.}
We augment the LLMs with \href{https://python.langchain.com/v0.2/docs/introduction/}{LangChain}. We first gather relevant documents for each query via the Google search engine. The top 10 URLs from the search are saved. The retrieved URLs then undergo a series of actions: (1) content retrieval using the \href{https://python.langchain.com/v0.2/docs/integrations/document_loaders/web_base}{WebBasedLoader}; (2) chunking to roughly 2000-character using the \href{https://api.python.langchain.com/en/latest/character/langchain_text_splitters.character.RecursiveCharacterTextSplitter.html}{RecursiveCharacterTextSplitter}, and (3) dense retrieval~\cite{DBLP:conf/emnlp/KarpukhinOMLWEC20} of the \emph{top-10} most relevant chunks based on cosine similarity on embeddings using the gte-Qwen2-1.5B-instruct~\cite{DBLP:journals/corr/abs-2308-03281, qwen2023qwen2} embedder.
These selected document chunks, along with the question, are compiled into a comprehensive prompt, as depicted in~\autoref{tab:RAG-prompts}, which is then provided to LLMs for generating responses.

\noindent\textbf{ReAct.}
ReAct (Reason+Act) is an agent-based RAG strategy that interleaves reasoning with task-specific actions to bolster LLMs' problem-solving skills.
During this process, the LLM agent makes decisions for subsequent actions based on the outcomes of prior retrievals and reasoning. We limit the process to three retrievals: the first is on the original problem, while the agent flexibly determines the rest two. 
From each retrieval, the LLM retrieves the 3 most relevant document chunks, resulting in a total of \emph{at most 9} chunks.
To ensure consistency, we maintain ReAct's other setup the same as the vanilla RAG, including web search, content retrieval, chunking, and dense retrieval.

\input{Prompts/generation_prompts}

%% file: Prompts/generation_prompts.tex
\begin{table}[ht]
\fontsize{7.5}{7.5}\selectfont
\setlength{\tabcolsep}{2pt} 
\centering
\begin{threeparttable}
\begin{tabularx}{\linewidth}{lX}
\toprule
\textbf{Prompt} &Content\\
\midrule
$p_{\text{basic}}$
&Your task is to provide an answer to the question.\\ 
\midrule
$p_{\text{comprehensive}}$
&You need to answer a debatable question. Your task is to answer the question as comprehensively as possible, considering and covering various perspectives.\\ 
\midrule
$p_{\text{detailed}}$
&You need to answer a debatable question. Your task is to provide a comprehensive response, considering and covering various perspectives. In your answer, ensure you:\newline1. Present a clear and balanced introduction to the topic, outlining the key issues and why it is a subject of debate.\newline2. Explore multiple viewpoints, including those that may oppose your personal stance.\newline3. Use evidence and examples to support each viewpoint. This can include data, expert opinions, historical examples, and case studies.\newline4. Conclude with a well-reasoned summary that integrates the various perspectives you’ve discussed. Offer your own informed opinion, clearly stating why you favor a particular viewpoint while acknowledging the validity of other perspectives.\\ 
\bottomrule
\end{tabularx}
\end{threeparttable}
\caption{\label{tab:system-prompts}
System prompts that require the model to answer questions with varying degrees of granularity.
}
\vspace{-1em}
\end{table}

%% file: Appens/case_study.tex
\section{Qualitative Analysis and Case Study}
\label{sec: case-studies}

\subsection{Deficiencies in Advanced LLMs}
\label{subsec: case-study-shortcoming-gpt4o}
We take the case of GPT-4o~\cite{openai2024gpt4o} to investigate the lingering deficiencies of advanced LLMs, a case study is provided in \autoref{tab:case-study-4o-weakness}.
Our main findings are:
\begin{packeditemize}
    \item \textbf{Insufficient evidence and source citation.} In a real debate, debaters need to support their arguments with evidence~\cite{wright1998debate}. Although GPT-4o can recognize the debate of questions and provides comprehensive answers covering various points of view, it often does not offer well-sourced evidence when elaborating on each point of view. Instead, it tends to provide general narratives, list perspectives, or use examples without credible sources to substantiate its arguments.
    \item \textbf{High-quality RAG improves performance.} We find high-quality retrieved results can effectively supplement GPT-4o's responses with relevant evidence to support its viewpoints, enhancing its capacity to address debatable questions with more credibility.
\end{packeditemize}

\noindent\textbf{Potential solutions} to address insufficient argumentation, potential solutions could be:
\begin{packeditemize}
    \item \textbf{Enhance domain-specific knowledge.} The model requires deeper knowledge of conflicts~\cite{xu2024knowledge}, achievable through domain-specific pretraining and RAG with high-quality sources. 
    \item \textbf{Encourage generation with reference.} The model should enhance its ability to cite sources in responses. Adapting strategies from existing research can help: as discussed in~\cite{DBLP:conf/eacl/WellerMWLKD24}, the model can refer to its pre-training data for generating responses. Additionally, as outlined in~\cite{DBLP:conf/emnlp/GaoYYC23}, employing RAG during response generation enables direct incorporation of cited evidence.
\end{packeditemize}

\sethlcolor{myred!20}

\begin{table*}[ht]
\fontsize{9}{9}\selectfont
\setlength{\tabcolsep}{1pt} 
\centering
\begin{threeparttable}
\begin{tabularx}{\linewidth}{X}
\toprule
\textbf{Question}  Should the government provide a universal basic income?\\
\midrule
\textbf{Answer w/o RAG} The question of whether the government should provide a universal basic income (UBI) is a complex and multifaceted issue \ldots Here are some key arguments for and against UBI: \newline Arguments in Favor of UBI: 
1. **Poverty Reduction**: \hl{UBI can provide a safety net for all citizens, helping to lift people out of poverty and reduce income inequality. } \textcolor{myred}{(This paragraph lacks arguments with clear sources.)} 
\ldots \newline
Arguments Against UBI: 1. **Cost**: \hl{Implementing a UBI could be extremely expensive and may require significant tax increases or reallocation of existing resources, which could be politically challenging. } \textcolor{myred}{(This paragraph lacks arguments with clear sources.)} 
\ldots
 \\
 \midrule
 \textbf{Answer w/ RAG} The question of whether the government should provide a universal basic income (UBI) is complex and multifaceted \ldots **Pros of UBI**: 1. **Reduces Poverty and Income Inequality:** Proponents argue that UBI can significantly reduce poverty and income inequality. \hl{For example, Namibia's Basic Income Grant trial reduced household poverty rates from 76\% to 37\% in one year, and Brazil's Bolsa Fam \'{i}edlia program lowered the poverty rate to the lowest level in 40 years in just six months. } \textcolor{myred}{(This is an argument with a clear source.) } \ldots \newline **Cons of UBI**: 1. **Cost:** Implementing UBI can be extremely costly. \hl{Andrew Yang's proposal for a \$1,000-a-month Freedom Dividend would have cost \$2.8 trillion annually, which is about half of the federal budget. } \textcolor{myred}{(This is an argument with a clear source.) }  Funding such a program would require significant tax increases or cuts to other social programs \ldots \\
\bottomrule
\end{tabularx}
\end{threeparttable}
\caption{\label{tab:case-study-4o-weakness}
Comparing GPT-4o's responses with and without RAG shows that RAG bolsters its capacity to present well-sourced arguments.
}
\vspace{-2em}
\end{table*}

\subsection{Gaps between Weak and Strong LLMs}

\sethlcolor{myred!20}

Taking Qwen2 7B~\cite{qwen2023qwen2} as a case study in \autoref{tab:case-study-qwen-weakness}, we pinpoint three main deficiencies typically found in the responses of models with moderate capabilities:
\begin{packeditemize}
    \item \textbf{Poor structural organization.} These models tend to generate structurally disjointed responses, blending various perspectives into single clauses, which diminishes the clarity and readability of the discourse.
    \item \textbf{Limited perspectives.} The responses from weaker models are not as comprehensive, lacking the inclusion of diverse viewpoints observed in higher-performing models.
    \item \textbf{Perspective imbalance.} There is a noticeable lack of neutrality in the responses. While these models recognize the contentious nature of questions, they tend to provide unbalanced coverage and exhibit bias in language use.
\end{packeditemize}

\noindent\textbf{Potential solutions} to improve responses that are poorly organized, incomplete or imbalance \wrt{} perspectives could be:

\begin{packeditemize}
    \item \textbf{SFT for stylistic answer presentation.} The model could be trained to deliver answers in a structured manner, \eg, using lists of answers with bolded perspectives to improve clarity. This formatting can be achieved through Supervised Fine-Tuning (SFT).
    \item \textbf{Alignment training and prompting.} To enhance response quality in terms of perspective diversity and balanced narrative, Preference learning with pairwise preferences data can be leveraged~\cite{DBLP:conf/nips/RafailovSMMEF23}. Another approach is to utilize prompting techniques such as self-reflection~\cite{DBLP:conf/nips/ShinnCGNY23} and role-based perspective-taking~\cite{xu2024walking} to explicitly require the model for more well-rounded and neutral answers.
\end{packeditemize}

\begin{table*}[ht]
\fontsize{9}{9}\selectfont
\setlength{\tabcolsep}{1pt} 
\centering
\begin{threeparttable}
\begin{tabularx}{\linewidth}{X}
\toprule
\textbf{Question} Should gay marriage be legalized? \\
\midrule
\textbf{Answer} \ldots I can share that many countries around the world have legalized gay marriage \hl{based on principles of equality, civil rights, and individual freedom. } \textcolor{myred}{(This section does not clearly distinguish these perspectives.) } These legalizations are often supported by arguments that \hl{same-sex couples should have the same rights as heterosexual couples, including the right to marry and form a legally recognized union. } \textcolor{myred}{(This section overlooks issues like the mental health of minority groups.) } Opponents might argue for various reasons, \hl{such as traditional definitions of marriage, religious beliefs, or concerns about social impact. } \textcolor{myred}{(This section overlooks issues like the impact of same-sex marriage on child development and the slippery slope argument, and it disproportionately focuses on the supportive views without clearly distinguishing these perspectives.) } Ultimately, the decision to legalize gay marriage is typically made through \ldots\\
\midrule
\textbf{Question} Is peer pressure a good thing? \\
\midrule
\textbf{Answer} Peer pressure can have both positive and negative effects, depending on the situation and context. \hl{Peer pressure can lead to positive outcomes such as promoting better behavior, enhancing social skills, and adhering to norms, but these are often discussed together with the negatives like encouraging harmful behaviors, promoting conformity, and influencing self-esteem. } \textcolor{myred}{(This section does not clearly distinguish these perspectives.) } For instance, while \hl{peer pressure can encourage someone to study harder or exercise and develop better social skills, } \textcolor{myred}{(This section overlooks perspectives such as accountability.) } \hl{it is often not separated from the pressure to engage in risky behaviors like drinking or reckless driving and often combined with pressures that enforce strict adherence to group norms, potentially leading to a loss of individuality. } \textcolor{myred}{(This section overlooks issues such as self-awareness and critical thinking.) } \\
\bottomrule
\end{tabularx}
\end{threeparttable}
\caption{\label{tab:case-study-qwen-weakness}
Two examples that illustrate multiple deficiencies in Qwen2 7B's responses.
}
\vspace{-2em}
\end{table*}

%% file: Appens/interpretation.tex
\section{A Practical Interpretation of the P.D. Metric}
\label{sec:interpretation}

{\small

Given the difficulty in interpreting the P.D. metric's absolute value, we offer an interpretation of the P.D. value's change to signify the comparative quality between two model answers, $A_1, A_2$.
Given the following definitions:

\begin{itemize}
    \item Perspective Diversity (P.D.)\footnote{Here, we use $A$ to denote $\texttt{chatTemplate}(\texttt{concat}(A,\text{``Please restate.''}))$ in \autoref{eq: PD} for simplicity.}:
    \[
    \text{P.D.} = \sum_{i=1}^{n} \text{PPL}(\text{PA}^i | A)
    \]
    \item Conditional perplexity:
    \[
    \text{PPL}(Y|X) = \exp \left( -\frac{1}{N} \sum_{i=1}^{N} \log P(y_i | X, y_{<i}) \right)
    \]
    \item We denote the \textbf{Change in Probability} for partial answer $i$ as \(\Delta P^i\), assuming $\mid \text{PA}^i \mid = N$:
    \begin{equation}
    \Delta P^i = \frac{P(\text{PA}^i|A_2)}{P(\text{PA}^i|A_1)} = \prod_{j=1}^{N} \frac{P(\text{PA}^i_j | A_2, \text{PA}^i_{<j})}{P(\text{PA}^i_j | A_1, \text{PA}^i_{<j})}
    \end{equation}
\end{itemize}

The \textbf{Change in Perspective Diversity} (\(\Delta \text{P.D.}\)) is defined as:
\begin{equation}
    \Delta \text{P.D.} = \sum_{i=1}^{n} \text{PPL}(\text{PA}^i | A_2) - \sum_{i=1}^{n} \text{PPL}(\text{PA}^i | A_1) =\sum_{i=1}^{n} \left( \text{PPL}(\text{PA}^i | A_2) - \text{PPL}(\text{PA}^i | A_1) \right)    
\end{equation}

Substitute the formula of conditional perplexity, we get:

\[
\Delta \text{P.D.} = \sum_{i=1}^{n} \left( \exp \left( -\frac{1}{N} \sum_{j=1}^{N} \log P(\text{PA}^i_j | A_2, \text{PA}^i_{<j}) \right) - \exp \left( -\frac{1}{N} \sum_{j=1}^{N} \log P(\text{PA}^i_j | A_1, \text{PA}^i_{<j}) \right) \right)
\]

Applying the approximation \(\exp(x) \approx 1 + x\) for small \(x\), we get:

\begin{align*}
    \Delta \text{P.D.} &\approx \sum_{i=1}^{n} \left( \left( 1 - \frac{1}{N} \sum_{j=1}^{N} \log P(\text{PA}^i_j | A_2, \text{PA}^i_{<j}) \right) - \left( 1 - \frac{1}{N} \sum_{j=1}^{N} \log P(\text{PA}^i_j | A_1, \text{PA}^i_{<j}) \right) \right) \\
    & = \sum_{i=1}^{n} \left( - \frac{1}{N} \sum_{j=1}^{N} \log P(\text{PA}^i_j | A_2, \text{PA}^i_{<j}) + \frac{1}{N} \sum_{j=1}^{N} \log P(\text{PA}^i_j | A_1, \text{PA}^i_{<j}) \right)\\
    & = - \frac{1}{N} \sum_{i=1}^{n} \sum_{j=1}^{N} \log P(\text{PA}^i_j | A_2, \text{PA}^i_{<j}) + \frac{1}{N} \sum_{i=1}^{n} \sum_{j=1}^{N} \log P(\text{PA}^i_j | A_1, \text{PA}^i_{<j})
\end{align*}

Applying the approximation that all \(\Delta P^i, i = 1\cdots n\) are the same value for simplicity, we get:
\[
\Delta \text{P.D.} \approx \sum_{i=1}^{n} \log (\frac{1}{\Delta P^i})
\]

Thus:
\begin{equation}
\Delta P^1 \approx \exp (-\frac{\Delta \text{P.D.}}{n})
\label{eq:interpretation}
\end{equation}

\autoref{eq:interpretation} shows that the change in probability \(\Delta P^1\) can be approximated using the change in P.D. \(\Delta \text{P.D.}\), and the effect is exponential. 
For example, when taking $n =3$ (given the fact that the average number of partial answers in \dataset{} $\approx 3.7$) and $\Delta\text{P.D.}=-2$, it suggests that the approximated probability of generating one partial answer under $A_2$ is $1.95$ times higher than under $A_1$.
In other words, the backbone language model finds it nearly twice as easy to recover the partial answer from $A_2$ due to the reduction of 2 in the P.D. score. 
}